\documentclass{article}

\usepackage{arxiv}

\usepackage[utf8]{inputenc} 
\usepackage[T1]{fontenc}    
\usepackage{hyperref}       
\usepackage{url}            
\usepackage{booktabs}       
\usepackage{amsfonts}       
\usepackage{nicefrac}       
\usepackage{microtype}      
\usepackage{lipsum}
\usepackage{graphicx}
\usepackage{lineno,hyperref}
\usepackage{scalefnt}
\usepackage{multirow}
\usepackage{amsmath}
\usepackage{subcaption}

\title{COVID-19 identification in chest X-ray images\\ on flat and hierarchical classification scenarios}

\author{
  Rodolfo~M.~Pereira \\
  Instituto Federal do Paraná (IFPR) and\\
  Pontifícia Universidade Católica\\ do Paraná (PUCPR)\\
  Pinhais--Paraná, Brazil \\
  \texttt{rodolfomp123@gmail.com} \\
   \And
  Diego~Bertolini \\
  Universidade Tecnológica Federal\\ do Paraná (UTFPR)\\
  Campo Mourão--Paraná, Brazil\\
  \texttt{diegobertolini@gmail.com} \\
  \And
  Lucas~O.~Teixeira\\
  Departamento de Informática\\
  Universidade Estadual de Maringá (UEM)\\
  Maringá--Paraná, Brazil\\
  \texttt{lucasteixeira@gmail.com}\\
  \And
  Carlos~N.~Silla~Jr.\\
  Pontifícia Universidade Católica\\ do Paraná (PUCPR)\\
  Curitiba--Paraná, Brazil\\
  \texttt{carlos.sillajr@gmail.com} \\
  \And
  Yandre~M.~G.~Costa\\
  Departamento de Informática\\
  Universidade Estadual de Maringá (UEM)\\
  Maringá--Paraná, Brazil\\
  \texttt{yandre@din.uem.br} \\
}

\begin{document}
\maketitle

\begin{abstract}
The COVID-19 can cause severe pneumonia and is estimated to have a high impact on the healthcare system. The standard image diagnosis tests for pneumonia are chest X-ray (CXR) and computed tomography (CT) scan. CXR are useful in because it is cheaper, faster and more widespread than CT. This study aims to identify pneumonia caused by COVID-19 from other types and also healthy lungs using only CXR images. In order to achieve the objectives, we have proposed a classification schema considering the multi-class and hierarchical perspectives, since pneumonia can be structured as a hierarchy. Given the natural data imbalance in this domain, we also proposed the use of resampling algorithms in order to re-balance the classes distribution. Our classification schema extract features using some well-known texture descriptors and also using a pre-trained CNN model. We also explored early and late fusion techniques in order to leverage the strength of multiple texture descriptors and base classifiers at once. To evaluate the approach, we composed a database, named RYDLS-20, containing CXR images of pneumonia caused by different pathogens as well as CXR images of healthy lungs. The classes distribution follows a real-world scenario in which some pathogens are more common than others. The proposed approach achieved a macro-avg F1-Score of 0.65 using a multi-class approach and a F1-Score of 0.89 for the COVID-19 identification in the hierarchical classification scenario. As far as we know, we achieved the best nominal rate obtained for COVID-19 identification in an unbalanced environment with more than three classes. We must also highlight the novel proposed hierarchical classification approach for this task, which considers the types of pneumonia caused by the different pathogens and lead us to the best COVID-19 recognition rate obtained here.

\end{abstract}



\section{Introduction}

The most recent novel coronavirus, officially named Severe Acute Respiratory Syndrome Coronavirus 2 (SARS-CoV-2), causes the Coronavirus Disease 2019 (COVID-19) \cite{world2020coronavirus}. The COVID-19 can cause illness to the respiratory system, fever and cough and in some extreme cases can lead to severe pneumonia \cite{guan2020clinical}. Pneumonia is an infection that causes inflammation primarily in the lungs' air sacs responsible for the oxygen exchange \cite{guan2020clinical}. 

Pneumonia can be caused by other pathogens besides SARS-CoV-2, such as bacterias, fungi and other viruses. Several characteristics can influence its severity: weak or impaired immune system, chronic diseases like asthma or bronchitis, elderly people and smoking. The treatment depends on the organism responsible for the infection, but usually requires antibiotics, cough medicine, fever reducer and pain reliever. Depending on the symptoms, the patient may need to be hospitalized; in severe cases the patient must be admitted into an intensive care unit (ICU) to use a mechanical ventilator to help breathing \cite{musher2014community}.

The COVID-19 pandemic can be considered severe due to its high transmissibility and seriousness \cite{tolksdorf2020influenza}. The impact in the healthcare system is also high due to the amount of people that needs ICU admission and mechanical ventilators for long periods \cite{grasselli2020critical}. In this scenario, early diagnosis is crucial for correct treatment to possibly reduce the stress in the healthcare system. In this context, artificial intelligence (AI) based solutions can provide a cheap and accurate diagnosis for COVID-19 and other types of pneumonia.

The standard image diagnosis tests for pneumonia are chest X-ray (CXR) and computed tomography (CT) scan. The CXR is the primary radiographic exam to evaluate pneumonia, but it not as precise as the CT scan and has higher misdiagnosis rates. Nevertheless, the CXR is still useful because it is cheaper, faster, expose the patient to less radiation and is more widespread than CT scan \cite{self2013high, rubin2020role}. The task of pneumonia identification is not easy, the professional reviewing the CXR needs to look for white patches in the lungs, the white patches are the lungs' air sacs filled with pus or water. However, these white patches can also be confused with tuberculosis or bronchitis, for example.

In this study, we aim to explore the identification of different types of pneumonia caused by multiple pathogens using only CXR images. Despite the CT scan being the gold standard for the pneumonia diagnosis, we focused only on CXR images due to its reduced cost, fast result and its general availability, since the CT scan machines are still scarse and costly. Specifically, we considered pneumonia caused by viruses (COVID-19, SARS, MERS and Varicella), bacteria (Streptococcus) and fungi (Pneumocystis). Moreover, because of the recent pandemic which is ravaging the world, the main focus of this work is the COVID-19 pneumonia, and our principal goal is to reach the best possible identification rate for it among other types of pneumonia and healthy lungs. To support that, we have taken into account two perspectives in the results' evaluation: first, we considered all classes mentioned above and summarized the results using a macro-avg F1-Score; second, we considered only the COVID-19 class and summarized the results using F1-Score.

In order to achieve that, we composed a database, named RYDLS-20, using CXR images from the open source GitHub repository shared by Dr. Joseph Cohen\cite{cohen2020covid}, images from the Radiopedia encyclopedia\footnote{https://radiopaedia.org/articles/pneumonia} and healthy CXR images from the NIH dataset, also known as Chest X-ray14 \cite{wang2017chestx}. The distribution of classes reflect a real world scenario in which healthy cases are the majority, followed by viral pneumonia, bacterial and fungi pneumonia being the least frequent, in this order. The RYDLS-20 database was made available also as a contribution of this work.

Even though, our main goal is to identify COVID-19 pneumonia, we setup the problem considering two different scenarios. In the first one, we address it as a multi-class problem, aiming at classifying different types of pneumonia (i.e. flat classification). In this way, each CXR image has a single label associated with it. In the second scenario, we address the problem as a hierarchical classification problem, since we can structure the different kinds of pneumonia based on the kind of pathogens that caused it \cite{silla2011survey}. To leverage that, we conducted the analysis using both flat and hierarchical classification methods.

Furthermore, the CXR image dataset used in this work was built aiming to reflect the real world distribution of different types of CXR, in which some types of pneumonia are much more likely than others, and even pneumonia itself is less frequent than healthy cases. Thus, the database is very imbalanced. In this situation, it is common that the classification algorithm increases the likelihood of the most frequent classes and reduces the likelihood of the least frequent classes \cite{wang2012multiclass}. In order to deal with this, we applied some well-known resampling techniques in order to balance of the classes distribution.

By analyzing CXR images, we can observe that texture is one of the main visual attributes present in those images. So, we decided to extract features from CXR images by exploring some popular texture descriptors, and also a pre-trained CNN model, not to neglect the power of representation learning approaches. Thus, for the flat classification, using the extracted features, we applied some well-known multi-class classification algorithms. In parallel, we also applied a hierarchical classification approach on the same set of extracted features. It is worth mentioning that we also tried a pure deep learning (end-to-end CNN) approach, however the results were very bad, probably due to the small sample size and the class imbalance.

Since multiple features were extracted using different texture descriptions, we also experimented different fusion techniques to take advantage of each descriptor strength, both on early and late fusion modes \cite{snoek2005early}. In early fusion, the features extracted from different texture descriptors were combined before training and test. In late fusion, each set of features is trained individually, and the algorithm predictions are combined after the training.

The paper is organized as follows. Section \ref{sec:tehoretical} presents some theoretical background about the concepts used in the study. After that, Section \ref{sec:works} discuss some related works and how they were used in our problem, when it is the case. Subsequently, Section \ref{sec:method} details our proposed methodology. Section \ref{sec:experiments} presents details about our experimental setup. Section \ref{sec:results} describes the obtained results. Later, Section \ref{sec:discussion} presents a discussion on the obtained results. Finally, Section \ref{sec:conclusions} concludes the current study and describes some possibilities for future works.

\section{Theoretical Background}
\label{sec:tehoretical}

In this section we present concepts regarding the pneumonia disease and the COVID-19 pandemic, flat and hierarchical classification and where they converge in our work. Besides, we also present important background concerning data imbalanceness, how it impacts our work and how to deal with it.

\subsection{COVID-19 Pandemic and Pneumonia Disease}

The COVID-19 outbreak was first reported in Wuhan, China at the end of 2019, it spread quickly around the World in a matter of months. The evidence points to an exponential growth in the number of cases, as of right now there are more than 2 million confirmed cases worldwide \cite{world2020coronavirus}. 

The epidemiological characteristics of COVID-19 are still under heavy investigation. The evidence so far shows that approximately 80\% of patients are in mild conditions (some are even asymptomatic) and 20\% are in serious or critical conditions. Moreover, around 10\% need to be admitted into an ICU unit to use mechanical ventilators. The fatality rate seems to be 2\%, but some specialists estimated it to be lower around 0.5\% \cite{novel2020epidemiological,remuzzi2020covid}. The ICU admission is that main concern since there are a limited number of units available.

One of main complications caused by COVID-19 is pneumonia. Pneumonia is an infection of the portion of the lung responsible for the gas transfer (the alveoli, alveolar ducts and respiratory bronchioles), called pulmonary parenchyma, that can be caused by different organisms, such as viruses, bacteria or fungi. Pneumonia cannot be classified as a single disease, but rather as a group of different infections with different characteristics \cite{mackenzie2016definition}.

Given that, pneumonia is considered a group of diseases, the diagnosis for each is also different. However, radiologic images, such as CXR and CT scan, are commonly used as one of the first exams to diagnose pneumonia of any kind. This happens because all kinds of pneumonia causes inflammation in the lungs, and that is captured by the radiologic images \cite{o2014radiological}.

Both CXR and CT scan are radiologic images that can be used aiming at identifying the pneumonia inflammation. CT scan is considered the gold standard over CXR since it is more precise. However, it has some drawbacks: it is more expensive, slower to be obtained and to an extent still rare \cite{self2013high}. Some CT scan machines can cost up to millions of dollars, and X-ray machines cost roughly ten times less than that. So, there are still reasons to use CXR images to diagnosis pneumonia.

Following that, pneumonia detection in CXR images can be difficult even for experienced radiologists. The inflammation appear as white patches in the lungs, they can be vague, overlapped with other diseases (asthma for example) and can even be confused with benign patches. In this context, artificial intelligence solutions can be very useful to aid the diagnosis.

\subsection{Flat and Hierarchical Classification}

When we talk about flat classification we are referring to binary, multi-class and multi-label classification problems. While in the binary classification problems there are only two different classes, in the multi-class problems there are multiple classes, but with only one output per sample. Moreover, in the multi-label classification problems, each instance may be associated to one or more labels.

Considering the described classification contexts, the problem of identifying types of pneumonia based on features extracted from CXR images can be naturally casted as multi-class classification problem, since we have one label associated to each sample. However, if we look at this same problem from another perspective, we may conclude that there is a hierarchy between the pathogens that causes pneumonia.

Hierarchical Classification is a particular type of classification problem, in which the output of the learning process is defined over a specific class taxonomy. According to Silla et al. \cite{silla2011survey}, this taxonomy can be considered a structured tree hierarchy defined over a partially order set (\textit{C}, $\prec$), where \textit{C} is a finite set that enumerates all class concepts in the application domain, and the relation $\prec$ represents a ``IS-A'' relationship. 

\begin{figure}[hbtp]
    \centering
    \includegraphics[width=0.7\textwidth]{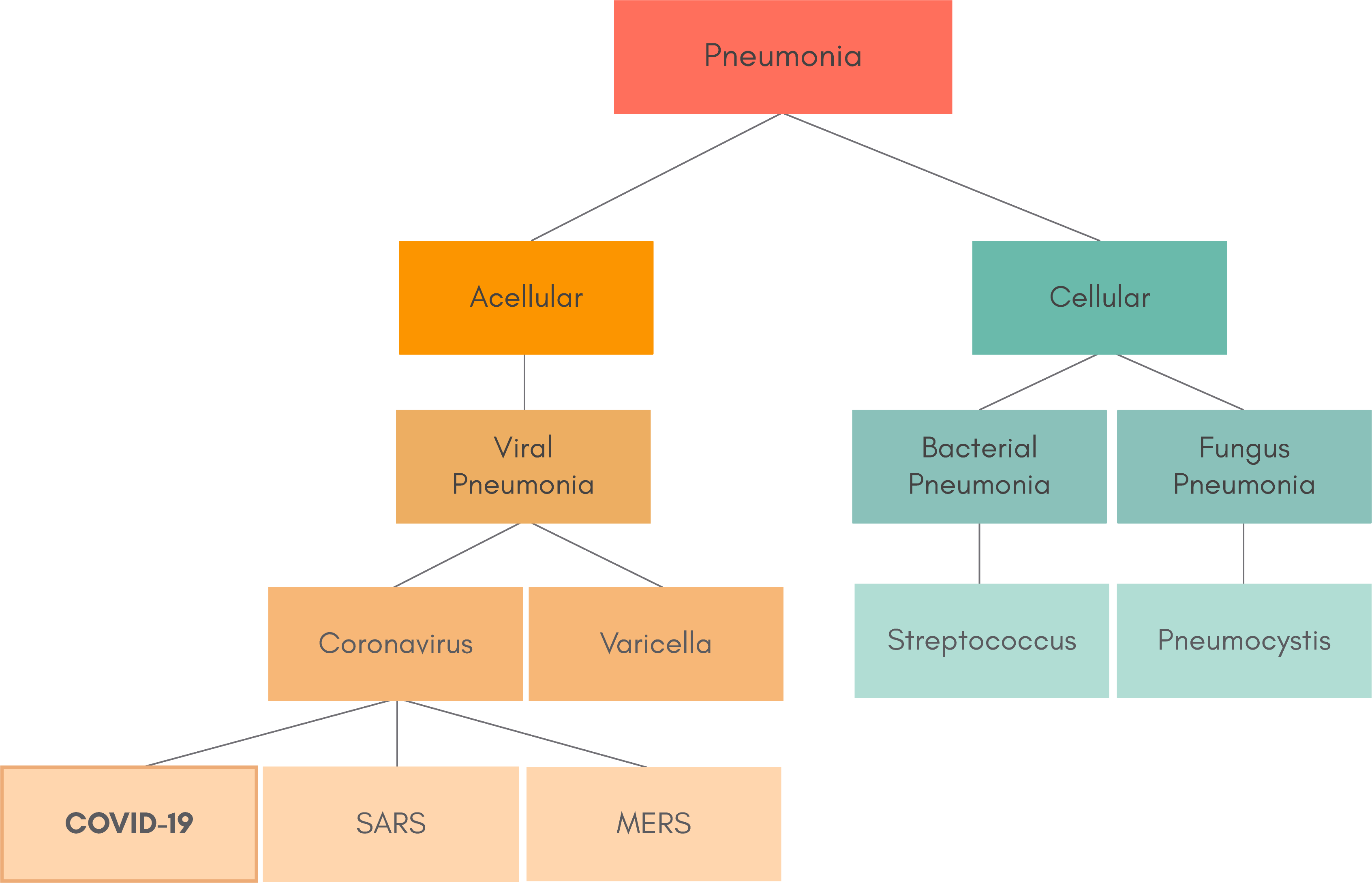}
    \caption{The hierarchical class structure of pneumonia caused by micro-organisms.}
    \label{hierarchy_pneumo}
\end{figure}

Figure \ref{hierarchy_pneumo} shows how the types of pneumonia caused by micro-organisms can be hierarchically organized. We may observe that there is a total of fourteen labels, in which seven are leaf nodes, i.e., which are the actual type of pneumonia. There are pneumonia caused by micro-organisms Acelullar/Virus and Cellular. By its turn, the Acelullar/Virus pneumonia can be subdivided into Coronavirus and Varicella, and the Celullar into Bacteria/Streptococcus and Fungus/Pneumocystis. Furthermore, the Coronavirus can be further subcategorized into COVID-19, SARS and MERS. This hierarchy is based on the structure developed in 10th revision of the International Statistical Classification of Diseases and Related Health Problems (ICD-10) \cite{american2019icd}.

According to \cite{silla2011survey}, there are different ways to tackle a hierarchical classification problem regarding the categorization of the classification process. In the Local Classifiers (LC) approach, the hierarchy is partially taken into account by using a local information perspective, creating binary/multi-class classifiers to deal with the problem in a local way. Moreover, in the Global Classifiers (GC) approach, a single classification model is built from the training data, taking into account the class hierarchy as a whole during the classification process. As we want the classifier to aim into the whole classes hierarchy in order to find important information between the pneumonia labels, in this paper we have used a GC approach to deal with the hierarchical classification problem.

\subsection{Imbalanceness Data and Resampling}

Many researchers face class imbalance distribution issues, mostly when working with real world datasets. Usually the classifiers are focused in the minimization of the global error rate and thus, when dealing with imbalanced datasets, the algorithms tend to benefit the most frequent classes (known as majority classes). Nevertheless, depending on the problem, the main interest of the task could be on properly labeling the rare patterns, i.e., the less frequent classes (known as minority classes), such as in credit card fraud detection \cite{kumar2015hierarchical} and medical image classification \cite{arias2016medical, bai2019nhl, abdulrazzaq2019xmiar}.

Following this reasoning, we may include the task being investigated in this work into the list of naturally imbalanced problems. In a real world scenario, the identification of types of pneumonia in CXR images may be also considered as an imbalanced learning task because there are more people with healthy lungs than people with lungs affected by the pneumonia disease. Moreover, regarding the number of people with the different types of pneumonia, there is also the imbalanceness factor. As the COVID-19 disease reached a pandemic status in 2020 \cite{world2020coronavirus}, the number of people with pneumonia caused by COVID-19 is much higher than the number of people with pneumonia caused by other pathogens such as SARS, MERS, Varicella, Streptococcus and Pneumocystis.

In order to deal with the imbalancess issue in classification datasets, several methods have been proposed in the literature \cite{fernandez2013analysing} and data level solutions are the most well-known and used techniques. The main objective of these techniques is to re-balance the classes distribution by resampling the dataset to diminish the effect of the class imbalanceness, i.e., preprocessing the dataset before the training phase.

\begin{figure}[htbp]
  \centering
    \includegraphics[width=0.6\columnwidth]{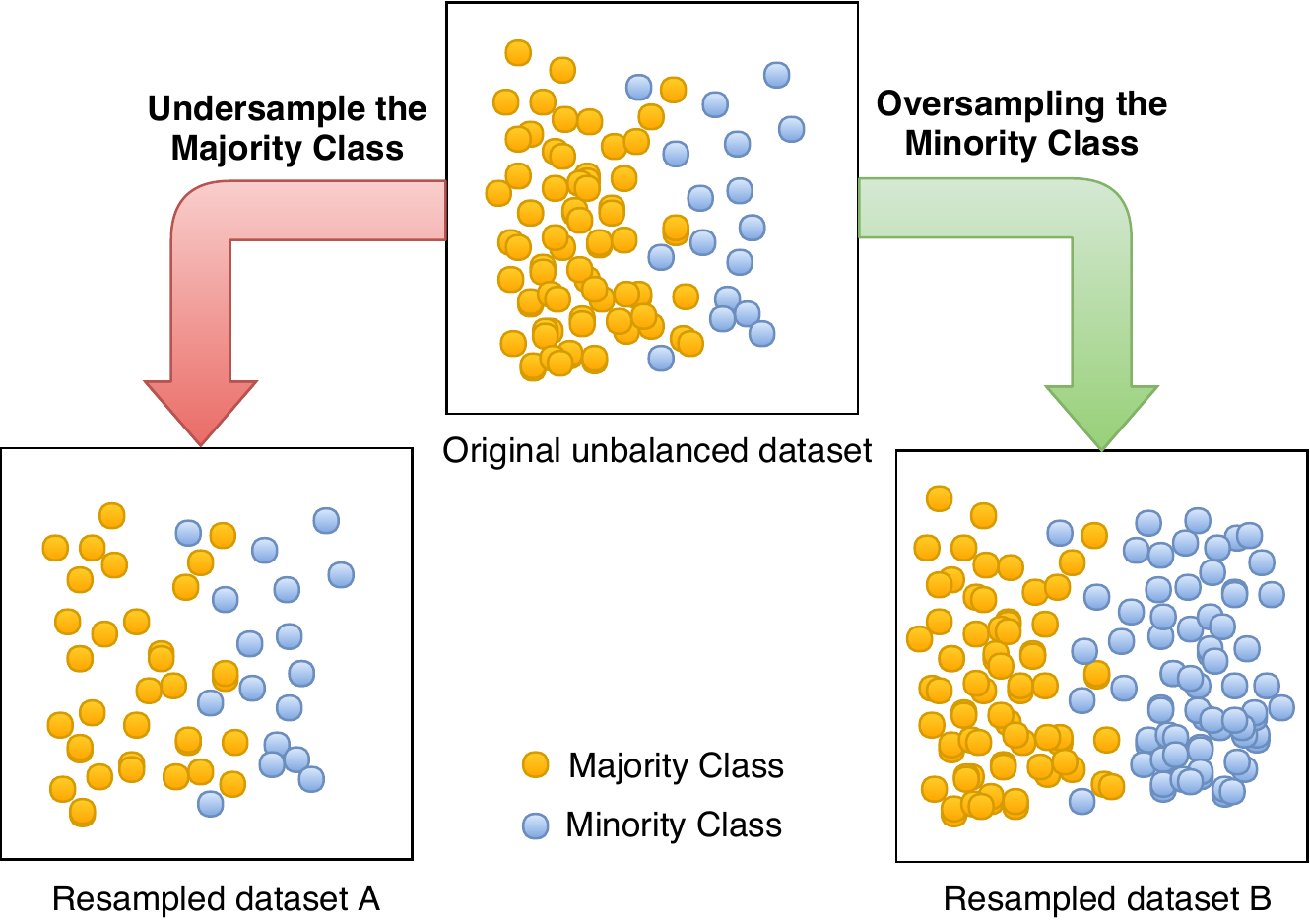}
    \caption{Different classes distribution in a binary labeled dataset.}
  \label{imb-schema}
\end{figure}

The resampling methods can be subdivided in two categories: oversampling and undersampling. Both are used to adjust the class distribution of a dataset, i.e., the ratio between the different classes in the dataset. While in an undersampling method some instances from the majority class are removed in order to balance the samples distribution, in an oversampling technique, some instances from the minority class are duplicated or synthetically created in order to balance the classes' distribution. Figure \ref{imb-schema} shows three different classes distribution in a typical binary dataset: (1) the original unbalanced dataset; (2) the resulting dataset after applying a undersampling over the majority class (Resampled dataset A); (3) the resulting dataset after applying a oversampling over the minority class (Resampled dataset B).

Although the resampling solutions were first defined and implemented for datasets with binary class distribution, we may apply them to multi-class imbalanceness problems as well. According to \cite{wang2012multiclass}, in order to apply these solutions in multi-class scenarios, most attention in the literature was devoted to class decomposition, i.e., the conversion of a multi-class problem into a set of binary class sub-problems. Two common decomposing schemas are: The One-Against-One (O-A-O) and the One-Against-All (O-A-A). While the O-A-O technique, first used in \cite{hastie1998classification}, proposes to train a classifier for each possible pair of classes, ignoring the examples that do not belong to the related classes, the O-A-A approach, introduced by \cite{rifkin2004defense}, builds a single classifier for each classe of the problem, considering the examples of the current class as positives and the remaining instances as negatives.

\begin{figure}[htbp]
\hspace{5.3cm}
\begin{subfigure}{.32\textwidth}
  \includegraphics[width=1\linewidth]{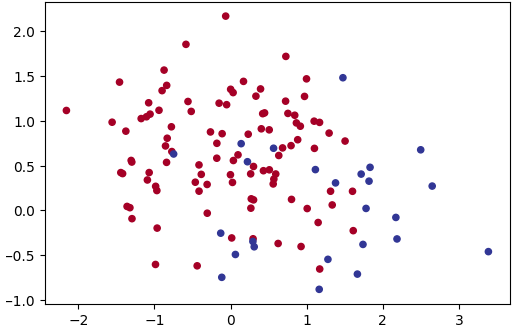} 
  \caption{Original.}
\end{subfigure}
\newline
\begin{subfigure}{.32\textwidth}
  \includegraphics[width=1\linewidth]{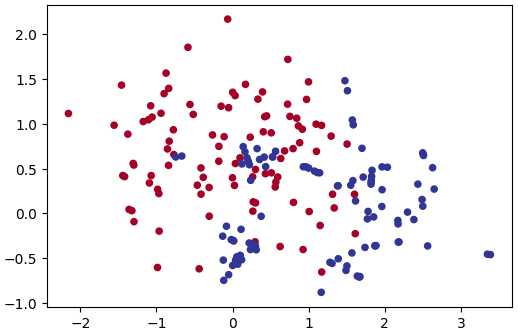}  
  \caption{SMOTE.}
\end{subfigure}
\begin{subfigure}{.32\textwidth}
  \includegraphics[width=1\linewidth]{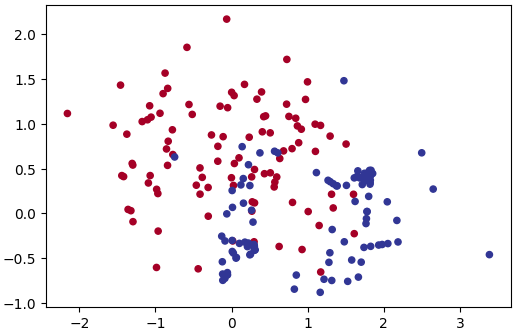}  
  \caption{SMOTE Bord-1.}
\end{subfigure}
\begin{subfigure}{.32\textwidth}
  \includegraphics[width=1\linewidth]{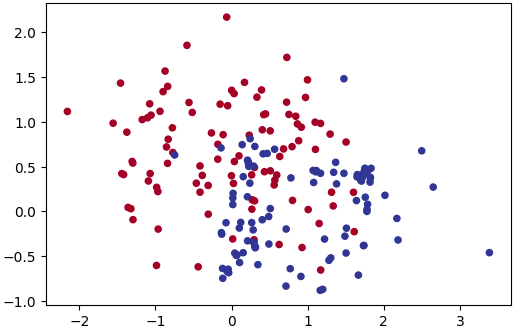}  
  \caption{SMOTE Bord-2.}
\end{subfigure}
\newline
\begin{subfigure}{.32\textwidth}
  \includegraphics[width=1\linewidth]{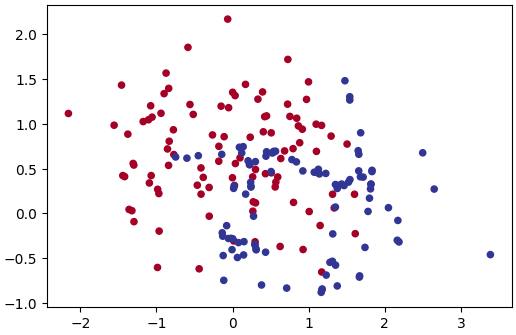}  
  \caption{ADASYN.}
\end{subfigure}
\begin{subfigure}{.32\textwidth}
  \includegraphics[width=1\linewidth]{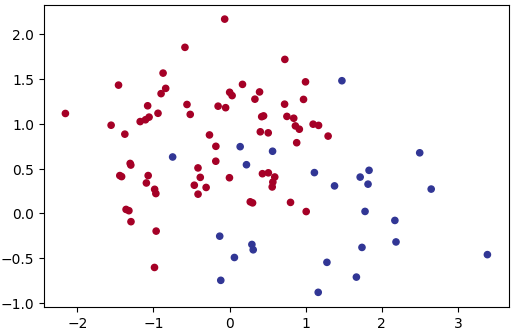}
  \caption{AllKNN.}
\end{subfigure}
\begin{subfigure}{.32\textwidth}
  \includegraphics[width=1\linewidth]{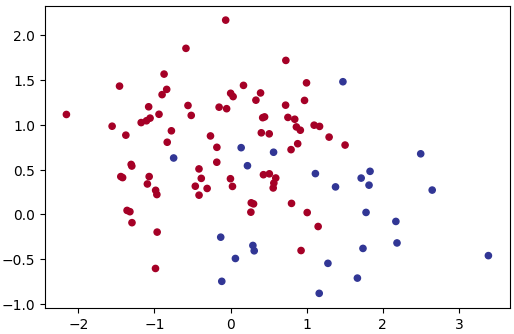}
  \caption{ENN.}
\end{subfigure}
\newline
\begin{subfigure}{.32\textwidth}
  \includegraphics[width=1\linewidth]{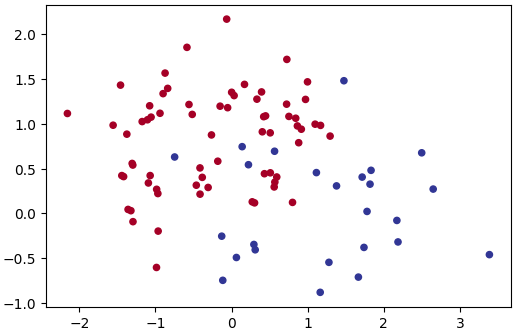}
  \caption{RENN.}
\end{subfigure}
\begin{subfigure}{.32\textwidth}
  \includegraphics[width=1\linewidth]{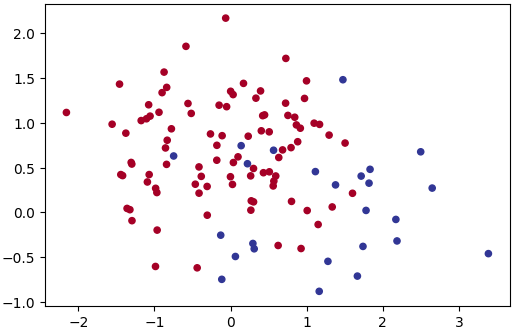}
  \caption{Tomek Link.}
\end{subfigure}
\begin{subfigure}{.32\textwidth}
  \includegraphics[width=1\linewidth]{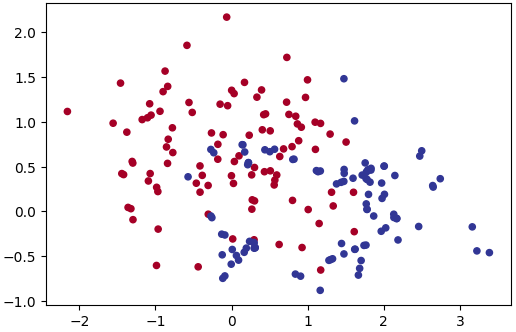}
  \caption{SMOTE+TL.}
\end{subfigure}
\caption{Example of datasets before and after applying the resampling techniques.}
\label{fig:res-exampl}
\end{figure}

In Table \ref{bres_alg} we present a brief review of the classic binary resampling algorithms, describing the main idea, strategy used, i.e., oversampling, undersampling or both (hybrid), and the paper in which it was proposed. In order to give a visual idea of the resampling techniques, Figure \ref{fig:res-exampl} shows an example dataset before and after applying the resampling methods. The example dataset is composed of two labels (blue and red) and two features (represented by axis y and x). While (a) shows the original datasets, (b)-(j) presents the resulting dataset after applying each one of the resampling techniques.

\begin{table}[htbp]
\scalefont{0.9}
\centering
\caption{Summary of Classic Binary Resampling Algorithms.}
\setlength{\tabcolsep}{3.5pt}
\begin{tabular}{lllc}
\hline
\multicolumn{1}{c}{Algorithm} & \multicolumn{1}{c}{Main Idea} & \multicolumn{1}{c}{Strategy} & \multicolumn{1}{c}{Reference} \\ \hline
ADASYN & \begin{tabular}[c]{@{}l@{}}Creates synthetic samples\\ for the minority class adaptively.\end{tabular} & Oversampling & \cite{he2008adasyn} \\
SMOTE & \begin{tabular}[c]{@{}l@{}}Creates synthetic samples \\ by combining the existing ones.\end{tabular} & Oversampling & \cite{chawla2002smote} \\
SMOTE-B1/B2 & \begin{tabular}[c]{@{}l@{}}Creates synthetic samples considering \\ the borderline between the classes.\end{tabular} & Oversampling & \cite{han2005borderline} \\
AllKNN & \begin{tabular}[c]{@{}l@{}}Removes samples in which a kNN\\ algorithm misclassifies them.\end{tabular} & Undersampling & \cite{tomek1976experiment} \\
ENN/RENN & \begin{tabular}[c]{@{}l@{}}Removes samples in which its label differs \\ from the most of its nearest neighbors.\end{tabular} & Undersampling & \cite{tomek1976experiment} \\
TomekLinks & \begin{tabular}[c]{@{}l@{}}Removes samples which are nearest \\ neighbors but has different labels.\end{tabular} & Undersampling & \cite{tomek1976experiment} \\
SMOTE+TL & Apply SMOTE and TomekLink algorithms. & Hybrid & \cite{batista2004study} \\ \hline
\end{tabular}
\label{bres_alg}
\end{table}

\section{Related Works}
\label{sec:works}

In this section, we describe some remarkable works presented in the literature which address one of the following topics, and that have directly influenced the development of this work: texture descriptors in medical images, pneumonia detection in CXR images, and COVID-19 pneumonia detection on CXR or CT scan images using artificial intelligence. The works will be described predominantly in chronological order, and we try to highlight the main facts related to each of them, such as: the feature extraction step, whether it is performed using handcrafted features or automated feature learning, classification model, database used in the experiments, type of image used in the experiments (CXR or CT), and the types of pneumonia investigated (COVID-19 among others).

The first work we are going to describe was presented by Nanni et al. \cite{nanni2010local} in 2010. In that work, the authors compared a series of handcrafted texture descriptors derived from the Local Binary Pattern (LBP), considering their use specifically in medical applications. The variants of LBP evaluated were Local Ternary Pattern (LTP), Eliptical Binary Pattern (EBP), and Elongated Quinary Pattern (EQP). These descriptors were evaluated on three different medical applications: i) pain expression detection, starting from facial images of the COPE database \cite{brahnam2007introduction}, taken from 26 neonates categorized into five different classes (i.e. rest, cry, air stimulus, friction, and pain); ii) Cell phenotype image classification using the 2D--Hela dataset \cite{chebira2007multiresolution}, a dataset composed of 862 single-cell images distributed into ten different classes (i.e. Actin Filaments, Endsome, Endoplasmic Reticulum, Golgi Giantin, Golgi GPP, Lysosome, Microtubules, Mitochondria, Nucleolus, and Nucleus); and iii) 
Pap smear (Papanikolaou test) aiming to diagnose cervical cancer. They used a dataset composed of 917 images collected at the Herlev University Hospital \cite{jantzen2005pap} using digital camera and microscope. The images were labeled according to seven different classes, being three of them related to normal states, and four of them related to abnormal states. After a comprehensive set of experiments using the Support Vector Machine (SVM) classifier, the authors verified that the EQP descriptor, or ensembles created using variations of EQP performed better for all the addressed tasks. For this reason, we decided to investigate the performance of EQP in the experiments accomplished in this work.

Still in 2010, Parveen and Sthik \cite{parveen2011detection} addressed pneumonia detection in CXR images. The authors suggested that the feature extraction could be properly made, at that time, by using Discrete Wavelet Transform (DWT), Wavelet Frame Transform (WFT), or Wavelet Packet Transform (WPT), followed by the use of fuzzy c-means classification learning algorithm. Looking backward, we can easily note that the feature extraction was still strongly coupled up to the handcrafted perspective. However, the efforts made by the authors aiming at find useful descriptors to properly capture the information about different kinds of lung infections is worth mentioning.

Scalco and Rizzi \cite{scalco2017texture} performed texture analysis of medical images for radiotherapy applications. In this sense, the authors applied texture analysis as a mathematical technique to describe the grey-level patterns of an image aiming at characterize tumour heterogeneity. They also carried out a review of the literature on texture analysis on the context of radiotherapy, particularly focusing on tumour and tissue response to radiations. In conclusion, the authors point out that texture analysis may help to improve the characterization of intratumour heterogeneity, favouring the prediction of the clinical outcome. Some important open issues concerning the interpretation of radiological images have been raised in that work. Among these issues, we can highlight the lack of a proper biological interpretation of the models that could predict the tissue response to radiation.

Although the COVID-19 outbreak is a quite recent event, it has been attracting a lot of attention from the society and also from the image analysis research community in particular, in view of the urgency of this matter. In this sense, Zhou et al. \cite{zhou2020improved} have just published a study describing a deep learning model for differentiating novel coronavirus pneumonia (NCP) and influenza pneumonia in chest computed tomography (CT) images. 
The work is one of the pioneers works that has brought to light some scientific evidences concerning the challenging pandemic which has been dramatically affecting the world. By this way, it can be taken as an important reference, mainly if we take into account that the study was developed by scientists from the country from where the outbreak has emerged.

The first point to be highlighted regarding that work is that, differently from the study presented in this work, Zhou et al. adopted CT images in their study. It is particularly important to emphasize this difference, because on the one hand, CT images are much better than CXR images due to its better capacity to show details from the pulmonary infection. On the other hand, CXR images are much cheaper and can be obtained in much less time, as already pointed in the introduction of this paper.

In the experimental protocol, Zhou et al. composed the training set using CT images scanned from 35 confirmed NCP patients, enrolled with 1,138 suspected patients. Among these images, it was included images from 361 confirmed viral pneumonia patients, being 156 of them influenza pneumonia patients. In summary, the study showed that most of the NCP lesions (96.6\%) are larger than 1 cm, and for 76.6\% of the lesions the intensity was below -500 Hu\footnote{Hounsfield unit, more details can be found in https://en.wikipedia.org/wiki/Hounsfield\_scale}, showing that these lesions have less consolidation than those provoked by influenza, whose nodes size ranges from 5 to 10 mm. Regarding the classification results, the deep model created obtained a rate above 0.93 for distinguishing between NCP and influenza considering the AUROC metric.

The authors also handle the transferability problem, aiming to avoid that the well-trained deep learning model performs poorly on data from unseen sources. In this way, Zhou et al. proposed a novel training schema, which they call Trinary schema. By this way, the model is supposed to better learn device independent features. The Trinary schema performed better than the Plain schema with specialists regarding the device-independence and consistence, achieving a F1 score of 0.847, while the Plain schema obtained 0.774, specialists 0.785, and residents 0.644.

Li et al. \cite{li2020artificial} also addressed COVID-19 identification on chest CT images using artificial intelligence techniques. For this purpose, the authors used a database composed of CT images collected from COVID-19 patients, other viral pneumonia patients, and also from patients not diagnosed with pneumonia. The images were provided by six Chinese hospitals, and the created database is composed of 2,969 images on the training set, being 400 from COVID-19 patients, 1,396 from other viral pneumonia patients, and 1,173 from non-pneumonia patients. In addition, it was created an independent test set with images from 68 COVID-19 patients, 155 other viral pneumonia patients, and 130 non-pneumonia patients, totaling 353 CT images.

A 3D deep learning model, which the authors call COVNet, was created using the ResNet-50 \cite{he2016deep} as a backbone. The model is fed by a series of CT slices and generates a classification prediction for the CT image considering the following three classes: COVID-19, other viral pneumonia, and non-pneumonia. After experimentation, the authors reported an AUROC value of 0.96 obtained for COVID-19, and 0.95 for other viral pneumonia.

Narin et al. \cite{narin2020automatic} evaluated COVID-19 detection on CXR images using three different deep neural network models (i.e. ResNet50, Inception-V3, and InceptionResNetV2). The dataset was composed using fifty COVID-19 patients images taken from the open source GitHub repository shared by Dr. Joseph Cohen \cite{cohen2020covid}, and more fifty healthy patients images from Kaggle repository ``Chest X-Ray Images (Pneumonia)''\footnote{https://www.kaggle.com/paultimothymooney/chest-xray-pneumonia}. The results were obtained using five-fold cross validation, and they are as follows: 98\% of accuracy using the ResNet50 model, 97\% of accuracy using the Inception-V3 model, and 87\% of accuracy for Inception-ResNetV2.

Gozes et al. \cite{gozes2020rapid} addressed COVID-19 detection and patient monitoring using deep learning models on CT images. By patient monitoring, the authors mean the evolution of the disease in each patient over time using a 3D volume, 
generating what they call ``Corona score''. The authors claim that the work is the first one developed to detect, characterize and track the progression of COVID-19. The study was developed using images taken from 157 patients, from China and USA. The authors make use of robust 2D and 3D deep learning models, they also modified and adapted existing AI models, combining the results with clinical understanding. The classification results, aiming at differentiate coronavirus images vs. non-coronavirus images obtained 0.996 of AUROC. Gozes et al. also claim that they successfully performed quantification and tracking of the disease burden.

Wang and Wong \cite{wang2020covid} created the COVID-Net, an open source deep neural network specially created aiming to detect COVID-19 on chest radiography images. To accomplish that, the authors curated the COVIDx, a dataset created exclusively to support the COVID-Net experimentation. The dataset is composed of 16,756 chest radiography images from 13,645 different patients taken from two distinct repositories. The authors describe in details the COVID-Net architecture design, and they also explain how one can get the dataset.

The initial network design prototype was created based on human-driven design principles and best practices, combined with machine-driven design exploration to produce the network architecture. 
The authors claim that the developed model obtained a good trade off between accuracy and computational complexity. In terms of recognition performance, they obtained 92.4\% of accuracy for the COVIDx test dataset as a whole. They also reported the following sensitivity rate for each kind of infection/non-infection image: 95\% for ``normal'' patients, 91\% for non-COVID-19 infection, and 80\% for COVID-19 infection. More details regarding the results, the created model, and the dataset can be found in \cite{wang2020covid}.

Khan \textit{et al.} \cite{khan2020coronet} designed the CoroNet, a Convolutional Neural Network (CNN) for detection of COVID-19 from CXR images. The CNN model is based in Xception (Extreme Inception) and contains 71 layers trained on the ImageNet dataset. The author also developed a balanced dataset to support and test their neural network configuration, which is composed of 310 normal, 330 bacterial, 327 viral and 284 COVID-19 resized CXR images. According to the authors, the proposed CoroNet achieved an average accuracy of 0.87 and a F1-Score of 0.93 for the COVID-19 identification. We can highlight the following main differences between their work and ours: (i) Their dataset do not consider an unbalanced realistic scenario, thus they do not use resampling techniques; (ii) Their dataset have only four classes and it is not publicly available for download; (iii) They did not use both handcraft and representation learning features. (iv) They did not investigate a hierarchical classification approach.

Ozturk et al. \cite{ozturk2020automated} proposed a deep model for early detection of COVID-19 cases using X-ray images. 
The author accomplished the classification both in binary (COVID vs. No-findings) and multi-class (COVID vs. No-Findings vs. Pneumonia) modes. The created model achieved an accuracy of 98.08\% for binary classes and 87.02\% for multi-class cases. The model setup was built the DarkNet model as a classifier YOLO object detection system. The authors made the codes available and they claim that the it can be used to create a tool to assist radiologists in validating their initial screening.

It is important to mention that as the identification of COVID-19 in CXR images is a hot topic nowadays due to the growing pandemic, it is unfeasible to represent the real state-of-the-art for this task, since there are new works emerging every day. However, we may observe that most of these works are aiming to investigate configurations for Deep Neural Networks, which is already somehow different from our proposal.

Table \ref{tab:works} presents a summary of the studies described in this section, focusing on their most important characteristics.
The main purpose of this table is to provide a practical way to find some important information regarding those works at a glance.

\begin{table}[htpb!]
\centering
\caption{Summary of the works described in this section.}
\label{tab:works}
\setlength{\tabcolsep}{3pt}
\begin{tabular}{lccc}
\hline
\multicolumn{1}{c}{Reference} & \multicolumn{1}{c}{Image Type} & \multicolumn{1}{c}{Database/applications} & \multicolumn{1}{c}{\begin{tabular}[c]{@{}c@{}}Computational/ML$^*$\\techniques \end{tabular}} \\ \hline
Nanni et al. \cite{nanni2010local} & \begin{tabular}[c]{@{}c@{}}Neonatal facial,\\fluorescence\\ microscope\\ and smear\\ cells images \end{tabular} & \begin{tabular}[c]{@{}c@{}}Three databases:\\Neonatal facial images,\\2D-HeLa dataset and\\ Pap smear datasets \end{tabular} & \begin{tabular}[c]{@{}c@{}}LBP, LPQ, EQP, \\LTP, EBP, ILBP\\ CSLBP and SVM\end{tabular} \\ \hline
Parveen and Sthik \cite{parveen2011detection}& CXR & Pneumonia detection & \begin{tabular}[c]{@{}c@{}}DWT, WFT, WPT \\and fuzzy C-means\\ clustering\end{tabular} \\ \hline
Scalco and Rizzi \cite{scalco2017texture}&\begin{tabular}[c]{@{}c@{}}CT, PET\\ and MR\end{tabular} & \begin{tabular}[c]{@{}c@{}}Tumour heterogeneity\\ characterization\end{tabular} &\begin{tabular}[c]{@{}c@{}}Grey-level\\ histogram, GLCM,\\ NGTDM, GLRLM\\and GLSZM\end{tabular}  \\ \hline
Zhou et al. \cite{zhou2020improved}& CT &\begin{tabular}[c]{@{}c@{}}NCP/influenza\\ differentiation images \\from 1,138 suspected \\patients, being 361 \\viral pneumonia, \\ 35 confirmed NCP and \\ 156 confirmed influenza \end{tabular}  & \begin{tabular}[c]{@{}c@{}}YOLOv3,\\VGGNet\\and AlexNet \end{tabular} \\ \hline
Li et al. \cite{li2020artificial}& CT & \begin{tabular}[c]{@{}c@{}}2,969 images obtained\\in Chinese hospitals\\400 NCP images \\ 1,396 other viral pneumonia \\ and 1,173 non-pneumonia \end{tabular} & \begin{tabular}[c]{@{}c@{}}COVNet\\deep learning\\model based on\\ResNet-50 \end{tabular} \\ \hline
Narin et al. \cite{narin2020automatic}& CT &\begin{tabular}[c]{@{}c@{}}NCP identification on a\\ dataset composed of\\ x-ray images from 50\\ healthy patients and\\ 50 COVID-19 patients \end{tabular}  & \begin{tabular}[c]{@{}c@{}}ResNet50,\\InceptionV3 and\\Inception-ResNetV2 \end{tabular} \\ \hline
Gozes et al. \cite{gozes2020rapid}& CT & \begin{tabular}[c]{@{}c@{}}NCP detection\\and analysis \\using images taken\\ from 157 patients \end{tabular} & \begin{tabular}[c]{@{}c@{}}2D and 3D deep\\learning models, \\and other\\AI models \end{tabular} \\ \hline
Wang and Wong \cite{wang2020covid}& CXR & \begin{tabular}[c]{@{}c@{}}NCP detection\\using 16,756 images\\taken from\\ 13,645 patients \end{tabular} & \begin{tabular}[c]{@{}c@{}}COVID-Net\\ a deep neural\\ network created\\ to detect NCP \end{tabular} \\ \hline
Khan \textit{et al.} \cite{khan2020coronet}& CXR & \begin{tabular}[c]{@{}c@{}}NCP detection\\using 1,251 images\\from four classes \end{tabular} & \begin{tabular}[c]{@{}c@{}}CoroNet\\ a CNN created\\ to detect NCP \end{tabular} \\ \hline
Ozturk \textit{et al.} \cite{ozturk2020automated}& CXR & \begin{tabular}[c]{@{}c@{}}NCP detection using\\ 500 pneumonia images\\ and 500 non-pneumonia images\end{tabular} & \begin{tabular}[c]{@{}c@{}}DarkNet\\ and YOLO \end{tabular} \\ \hline
$^*$ Machine Learning
\end{tabular}
\end{table}

\section{Proposed Method}
\label{sec:method}

As aforementioned in this paper, we focus on exploring data from CXR images considering different feature extraction methods to classify the different types of pneumonia and, consequently, identify COVID-19 among pneumonia caused by other micro-organisms. Thus, we chose specific approaches that could lead us to obtain the best benefit in terms of the classification performance for these specific classes. 

\begin{figure}[hbtp]
    \centering
    \includegraphics[width=1\textwidth]{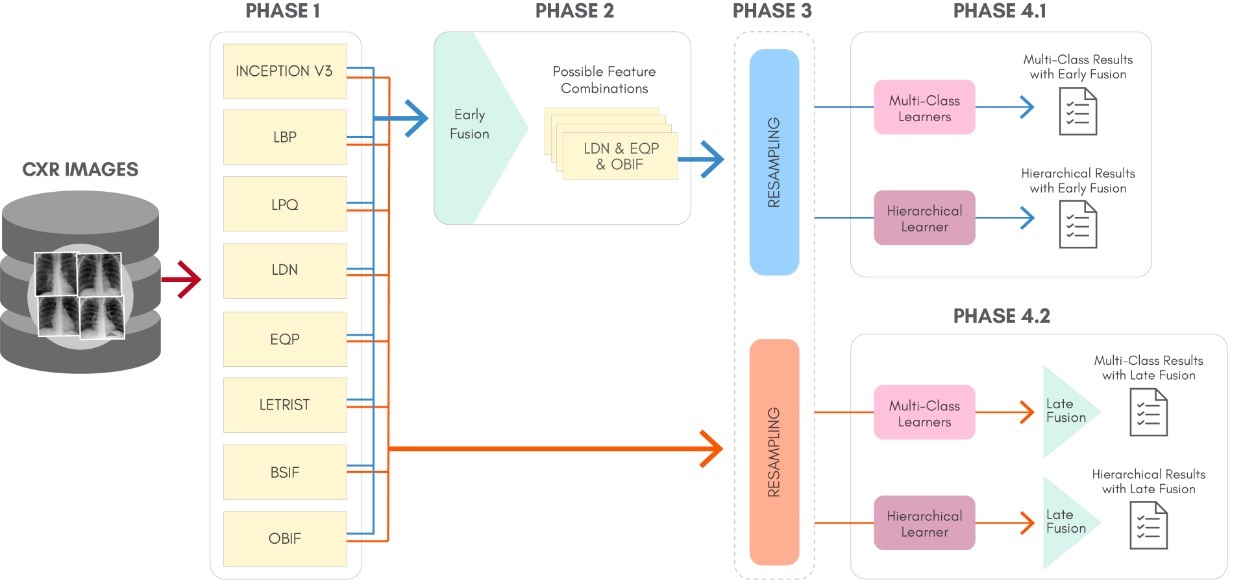}
    \caption{The proposed classification schema for the COVID-19 identification in CXR images.}
    \label{method}
\end{figure}

To better understand the proposal of this work, Figure \ref{method} shows a general overview for the classification approach, considering: The feature extraction process (Phase 1), the Early Fusion technique (Phase 2), the data resampling (Phase 3), the classification and generation of results for the multi-class and hierarchical scenarios (Phase 4.1 for Early Fusion and Phase 4.2 for Late Fusion). In Figure \ref{method}, Phase 3 is circulated with a dashed rounded rectangle because it is an optional phase, since the original features without resampling can also be used to generate the predictions.

It is important to inform that our method does not involve an automatic preprocessing step aiming to standardize the images before the feature extraction, although a manual crop to exclude undesirable patterns has been carried out. Thus, we are dealing with CXR images with different sizes. More information concerning the dataset will be described in section \ref{subsec:database}. In the following subsections we describe in details each one of the Phases present in Figure \ref{method}. 

\subsection{Feature Extraction (Phase 1)}

By analyzing CXR images, we can observe that texture is probably the main visual attribute we can find inside them. Thus, we have explored some well-successful texture descriptors described in the literature, both considering handcrafted and non-handcrafted approaches.

In the last decade, many researchers began using non-handcrafted features to describe patterns aiming to address classification problems. This kind of feature became a strong trend, thanks to the fact that they can be easily obtained, since it is not necessary to perform the feature engineering work, which in general is a laborious task and may require a refined knowledge regarding the problem classes, and also regarding the techniques that support handcrafted descriptors.

Even though the non-handcrafted descriptors have some clear advantages, we should remember that handcrafted features have peculiarities that can make them still quite useful to properly address many classification tasks. One of these advantages lies in the fact that handcrafted features tend to capture the patterns related to the problem, so to say, in a more stable fashion, due to the fact that these techniques often work in a deterministic way. Moreover, a more precise understanding regarding the patterns captured by handcrafted features from the images tends to be more feasible than when non-handcrafted features are used.

Anyway, in this work we moved efforts towards the employment of both these categories for feature extraction. By this way, we can evaluate both separately and, moreover, we also carried out several experimental setups performing a combination of them. In this sense, we can take advantage from the complementarity that may exist between both strategies used to obtain the descriptors, once they not necessarily make the same errors on the accomplishment of a given classification task, as shown in \cite{nanni2017handcrafted} and \cite{costa2017evaluation}.

In this section, we briefly describe the descriptors used in this work. The texture descriptors selected were chosen either because they presented good performance in general applications, or specifically in medical image analysis related applications.

\subsubsection{Local Binary Pattern (LBP)}
\label{subsec:lbp}

Presented by Ojala et al. \cite{lbpDescriptor}, LBP is a powerful texture descriptor, successfully experimented in several different applications which involve texture classification. LBP is found by calculating a binary pattern for the local neighborhood for each pixel of the image. The pattern has one position to each neighbor $i$ involved in the calculation, and it is calculated by subtracting the values of gray intensity of the central pixel ($g_c$) from the gray intensity of each neighbor ($g_i$) to get a distance $d$, such a way that if $d$ is greater or equal to zero, that position in the binary pattern assumes the value 1, otherwise it assumes 0, like shown in equation \ref{eq:lbp}.

\begin{equation}
\label{eq:lbp}
d=
\begin{cases}
             1, &  if\ g_i - g_c \geq 0\\
             0, & otherwise.\\
  \end{cases}
\end{equation}

The final texture descriptor for the image corresponds to a histogram which counts the occurrences of the binary patterns along the image. The number of possible binary patterns varies according to the setup previously defined to get pixels from the neighborhood, considering the neighborhood topology and number of neighbors for example. Regardless these details, this texture descriptor has been presenting impressive results on several different application domains for more than one decade \cite{bertolini2013texture, costa2012music, paula2014forest}. The details about how the patterns can be defined, and how we can set the parameters of the neighborhood can be found in \cite{lbpDescriptor}.

\subsubsection{Elongated Quinary Patterns (EQP)}
\label{subsec:eqp}
Elongated Quinary Pattern (EQP) \cite{nanni2010local} is basically a variation of LBP and LTP descriptors. The great difference from EQP to LBP and LTP descriptors is that the EQP uses a quinary pattern, not binary or ternary encoding, like LBP and LTP respectively. From a given grayscale image, let us denote \textbf{x} as the central pixel using a given topology, and \textbf{u} as the gray value of the neighboring pixels. In the EQP descriptor, we can assume five different values, instead of two or three, as proposed in the LBP and LTP respectively. Thus, in quinary encoding the difference $d$ is encoded using five values according to two thresholds $\tau_{1}$ and $\tau_{2}$, as described in Equation \ref{eq:eqp}.

\begin{equation}
\label{eq:eqp}
d=
\begin{cases}
             2, &  \textbf{u} \geq \textbf{x} + \tau  \\
             1, &  \textbf{x} + \tau_{1}  \leq \textbf{u}   <  \textbf{x} + \tau_{2}\\
             0, &  \textbf{x} - \tau_{1}  \geq \textbf{u}   <  \textbf{x} + \tau_{1} \\
             -1, & \textbf{x} - \tau_{2}  \leq \textbf{u}   <  \textbf{x} - \tau_{1} \\
             -2, & otherwise. \\
  \end{cases}
\end{equation}

The creators of EQP evaluated different topology patterns for the neighborhood of the central pixel, instead of considering only the circular neighborhood, as originally done for LBP and LTP. The elliptical topology showed better results. Moreover, the EQP proved to be a robust descriptor for different medical image problems, presenting results superior to other descriptors.

\subsubsection{Local Directional Number (LDN)}

LDN was originally experimented for the assessment of the face recognition task \cite{ldnDescriptor}. This technique tries to capture the texture structure, by encoding the information about the texture direction in a compact way. In this sense, this descriptor is supposed to be more discriminative than other methods. The directional information is obtained by using a compass mask and the prominent direction indices are used to encode this information. Sign-which is also used to distinguish between similar pattern with different intensity transitions. More details about this descriptor can be find in \cite{ldnDescriptor}.

\subsubsection{Locally Encoded Transform Feature Histogram (LETRIST)}

The LETRIST descriptor, proposed by Song et al. \cite{letristDescriptor} aims to present a simple and efficient texture descriptor. The authors describe some important characteristics that all texture descriptor must have: Discriminative, Invariant, Intense, Low-dimensional and Efficient. Basically, the histogram which corresponds to the LETRIST descriptor is based on the across feature and scale space of the image. The authors describe three main steps for generating the representation histogram. In the first step, from multiple input image scales and using Directional Gaussian Derivative Filters, extremum responses are computed. These extremum responses are used to feed linear and non-linear operators to quantitatively construct a set of transform features. This characterizes structures of local textures and their correlations with the input image. In the second step, the quantization is performed using various levels or binary threshold schemas. This step aims at greater robustness in terms of changes in lighting and rotation. Finally, in the third step, a joint cross-scale coding schema is carried out. In this way, it is possible to add discrete texture codes in a compact histogram representation. The authors describe LETRIST as a robust descriptor for different texture classification tasks \cite{letristDescriptor}. We can also point out that it is robust to Gaussian noise, changes in scale, rotation and lighting. In this way, LETRIST becomes a very interesting texture descriptor.

\subsubsection{Binarized Statistical Image Features (BSIF)}

Proposed by Kannala and Rahtu \cite{bsifDescriptor}, the BSIF texture descriptor was initially proposed for texture classification particularly on face recognition tasks. The BSIF descriptor is based both on LBP and LPQ descriptors. However, the authors emphasize that BSIF uses a schema based on statistics of natural images and not on heuristics, such as the descriptors LBP and LPQ. That is, from a small set of samples of natural images, the descriptor learns a fixed set of filters using Independent Component Analysis (ICA). For the generation of BSIF descriptors, the value of each pixel from an input image $M \times N$ is transformed into a binary string. 
In this work, the feature vectors generated using BSIF have 56 dimensions. An interesting study evaluating the robustness of 27 descriptors in palmprint recognition \cite{palmprint} describes that the BSIF descriptor was among the Top-3 best descriptors evaluated.

\subsubsection{Local Phase Quantization (LPQ)}

Proposed by Ojansivu and Heikkil{\"a} \cite{lpqDescriptor}, LPQ was originally proposed aiming to provide a good texture description for noised images, affected by blur. However, surprisingly LPQ has shown to be quite effective also to describe the textural content even for images not affected by blur. This descriptor is constructed by taking the coefficients that reveal the blur intensity of the image. It is done by using the phase of 2D Short Term Fourier Transform (STFT) over a window with a previously defined size, which is slid over the image. The mathematical details regarding the LPQ can be obtained in \cite{lpqDescriptor}.

\subsubsection{Oriented Basic Image Features (oBIFs)}

The BIF descriptor was originally designed for texture classification \cite{obifDescriptor}, but it also performs well in other tasks \cite{obifWriter}. Gattal et al. \cite{obifDescriptor2} proposed an extension of the BIF descriptor. The main idea is to categorize each location in the image into one of seven possible local symmetry classes. These types of local classes are the following: flat, slope, dark rotational, light rotational, dark line on light, light line on dark or saddle-like. To categorize each part of the image, six Derivative-of-Gaussian filters are used,  which is determined by the $\alpha$ parameter. The parameter $\varepsilon$ classifies the location probability as flat. The feature vector generated through the oBIFs descriptor has 23 dimensions. The orientations were quantified at four levels $(n = 4)$. Newell and Griffin \cite{obifDescriptor2,obifWriter} propose a change in the oBIFs descriptor aiming to improve its performance. In this sense, the rationale is that from two different oBIFs descriptors (using different parameters $\sigma$ and $\varepsilon$), it is possible to produce oBIFs column features with ${(5n+2)}^{2}$ dimensions. Thus, the number of dimensions was increased from 23 to 484.

\subsubsection{Automatically Learned Features with Inception-V3}

In the non-handcrafted scenario, we used the Inception-V3 \cite{szegedy2016rethinking} to perform feature extraction. This architecture proved to be more robust than other deep architectures, presenting low error rates in the ILSVRC-2012 challenge\footnote{http://image-net.org/challenges/LSVRC/2012/}. It also also presented results better than previous architectures, such as GoogleLeNet \cite{szegedy2015going}, PReLU \cite{he2016deep}, and VGG \cite{simonyan2014very}.
We have used zero padding to fill the images and keep their size in the standard. After the training of the Inception-V3, we used the 2,048 weights values of the penultimate layer of the net as feature vector. Before extracting the features, we applied transfer learning using the weights of an Inception-V3 trained on the IMAGENET Dataset \cite{krizhevsky2012imagenet}.

All codes employed in this work can be found through links in their respective papers. In Table \ref{tab:features} we describe the dimensions of the feature vectors and also the values set to their main parameters, aiming to facilitate the reproducibility by other researchers.

\begin{table}[htbp]
\centering
\caption{Features dimensions and main parameters.}\label{tab:features}
\begin{tabular}{ccc}
\hline
\textit{Feature} & \textit{Parameters} & \textit{Dimensions} \\ \hline
LBP               &   $LBP_{8,2}$                                      & 59 \\
EQP               &   \textit{loci = ellipse}                         & 256 \\
LDN               &   $mSize = 3$; $mask = kirsch$; $\sigma = 0.5$     & 56  \\
LETRIST           &   $sigmaSet = {1,2,4}$; \textit{noNoise}          & 413 \\
BSIF              &   \textit{filter = ICAtextureFilters-11$\times$11-8bit}   & 256 \\
LPQ               &   $winSize = 7$                                     & 256  \\
oBIFs             &   $\alpha = {2,4}$; $\varepsilon = 0.001$           & 484  \\
Inception-V3      &   \textit{default parameters}                     & 2,048  \\

\hline
\end{tabular}
\end{table}

\subsection{Early Fusion (Phase 2)}

This fusion technique was first used in Snock \textit{et al.} \cite{snoek2005early} and its main idea is to group the different features as a unique set of features to feed the learner. Thus, the method generates a unique dataset with all the chosen characteristics together. In our method, as we are using eight different features, we have decided to use 2$\times$2 and 3$\times$3 combinations, which lead us to a total of eighty four different feature sets.

\subsection{Resampling (Phase 3)}

This phase is a silver bullet point in the proposed classification schema. As already described in the introduction and background of this work, we are dealing with a naturally imbalanced problem, since there are much more cases of people with healthy lungs than with pneumonia. The use of prediction schemas that does not take into account this imbalance issue usually leads to bad performances. Moreover, the main focus of this paper is to identify the COVID-19 pneumonia among pneumonia caused by other pathogens, thus, we are aiming to increase the prediction scores for a class belonging to the set of minority labels.

In this situation, data augmentation might not be a good idea, since it can increase the overall chance of overfitting to the training data. This may happen because the operations performed in the image, such as rotations, translations, distortions and so on, does not change the pneumonia inflammation white patches spots at all. That is the main reason we decided to use resampling techquines over data augmentation techquines.

In order to balance the classes distribution in the training sets, we have applied binary resampling algorithms, aided by the O-A-A approach, as we have multi-classes datasets instead of binary. For the hierarchical classification scenarios, we have used the same resampling strategy, considering each leaf node label path as an individual multi-class label.

Furthermore, as described in the beginning of this section, this is an optional phase in the proposed classification schema presented in Figure \ref{method}, as we can perform classification without resampling to analyse and also combine the predictions in order to reach for better performances.

\subsection{Classification Approaches and Late Fusion (Phase 4)}

As already cited in this work, given the explicit hierarchy between the pneumonia, which is caused by multiple pathogens, in the proposed COVID-19 identification schema we propose to perform flat and hierarchical classification, which we do in Phase 4.1 (generation of Early Fusion results) and 4.2 (generation of Late Fusion results).

In opposition to early fusion strategies, the Late Fusion technique combines the output of the learners \cite{snoek2005early}. In general, this combination is achieved by calculating a only prediction involving all the predicted scores.

According to Kittler \textit{et al.} \cite{kittler1998combining}, the Late Fusion may achieve promising results in scenarios in which there is complementarity between the outputs. In these cases, the classifiers do not make the same misclassification and thus, when combined, they can help each other to give the best label prediction.

Among the most used fusion strategies, we can highlight the rules 
introduced by Kittler \textit{et al.}\cite{kittler1998combining} and which were used in this work: 

\begin{itemize}
    \item Sum rule (SUM): Corresponds to the sum of the predictions probabilities provided by each classifier for each label. 
    \item Product rule (PROD): Corresponds to the product between the predictions probabilities provided by each classifier for each label.
    \item Voting rule (VOTE): We contabilize the votes of the classifiers in the each possible label (considering the higher probability prediction) and choose the label with the most votes.
\end{itemize}

\begin{figure}[hbtp]
    \centering
    \includegraphics[width=1\textwidth]{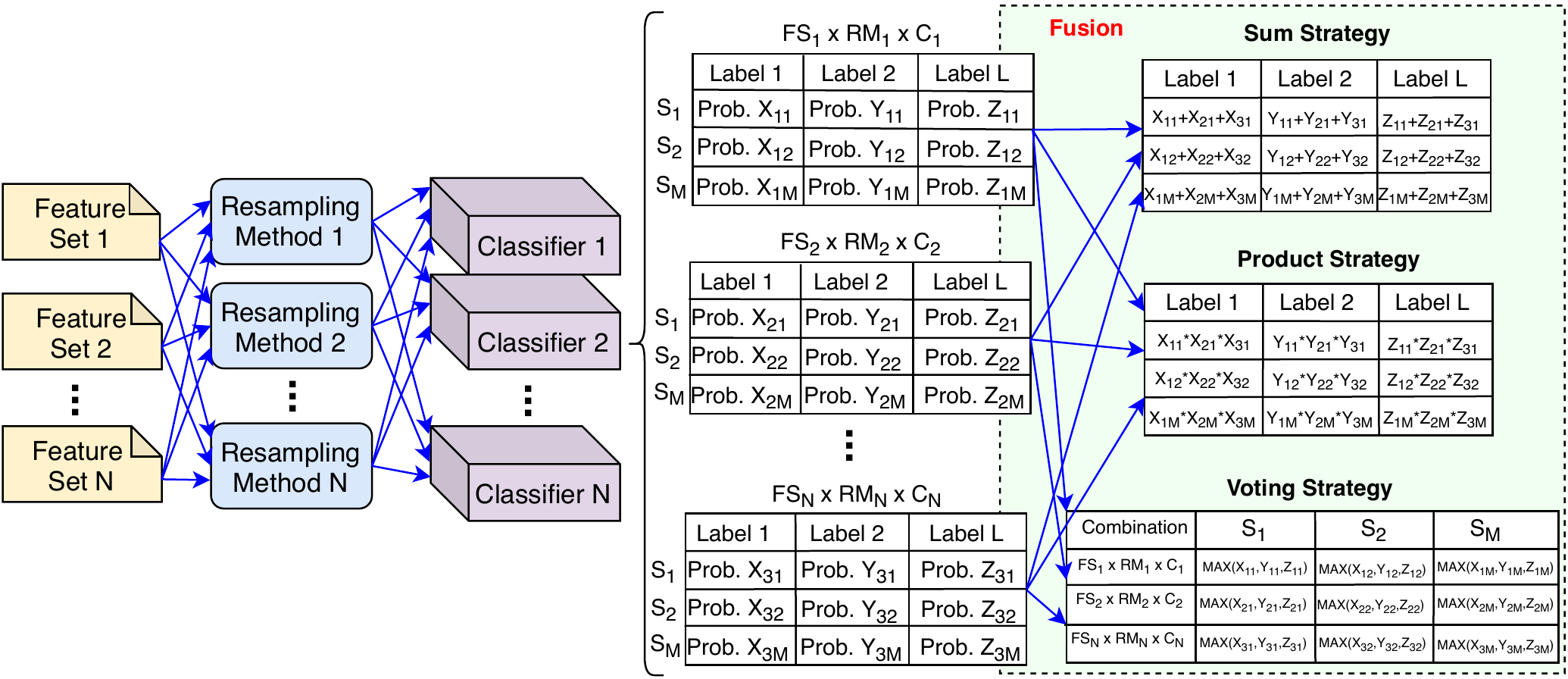}
    \caption{Example of combinations with late fusion strategies using the sum, product and voting strategies. The example dataset has M samples and L labels.}
    \label{late-fusion}
\end{figure}

Another aspect regarding the predictions integration is the criteria adopted to select the classifiers that will be used in the fusion. In this sense, we have tested the Top-N, Best-per-Feature and Best-per-Classifier fusion criteria. The Top-N consists of selecting the N tested scenarios (classifier + feature + resampling) with the best overall performance. The Best-on-Feature and Best-on-Classifier consists of using the best results for each feature and classifier, respectively. In our method, we have used N=5 in the Top-N approach. In the Best-on-Feature we have tested the combination 2$\times$3, 3$\times$3, 4$\times$4 and 5$\times$5 of the best result per feature, while in the Best-on-Classifier we have made the same combinations, but for the best classifiers. Figure \ref{late-fusion} presents an general schema of the late fusion technique used in this work. In the example we show details concerning how the probabilities are combined to generate the result.

\section{Experimental Setup}
\label{sec:experiments}

In this section we present the proposed database, algorithms, parameters and metrics used in this paper. It is important to observe that the database, as well as the experimental scripts used in this work are all freely available for download\footnote{https://drive.google.com/open?id=1J9I-pPtPfLRGHJ42or4pKO2QASHzLkkj}.

\subsection{The Database}
\label{subsec:database}

Table \ref{db-main-char} presented the main characteristics of the proposed database, which was named RYDLS-20. As it can be noted, the database is composed of 1,144 CXR images, with a 70/30 percentage of split between train/test. Moreover, there are seven labels, which can be further hierarchically organized into fourteen label paths.

\begin{table}[htbp]
\centering
\caption{RYDLS-20 main characteristics.}
\begin{tabular}{lc}
\hline
\textit{Characteristic} & \textit{Quantity} \\ \hline
Samples & 1,144 \\
Train & 802 \\
Test & 342 \\
Labels (Multi-Class Scenario) & 7 \\
Label Paths (Hierarchical Scenario) & 14 \\ \hline
\end{tabular}
\label{db-main-char}
\end{table}

The CXR images have diffrent sized and were obtaiend from three different sources:

\begin{itemize}
    \item COVID-19, SARS, Pneumocystis and Streptococcus images  were obtained in the open source GitHub repository shared by Dr. Joseph Cohen\cite{cohen2020covid}.
    \item Varicella and MERS images were obtained from the Radiopedia encyclopedia\footnote{https://radiopaedia.org/articles/pneumonia}.
    \item The Normal lung images were all obtained from NIH dataset, also known as Chest X-ray14 \cite{wang2017chestx}.
\end{itemize}

\begin{figure}[htbp]
\centering
\begin{subfigure}{.22\textwidth}
  \centering
  \includegraphics[width=.9\linewidth]{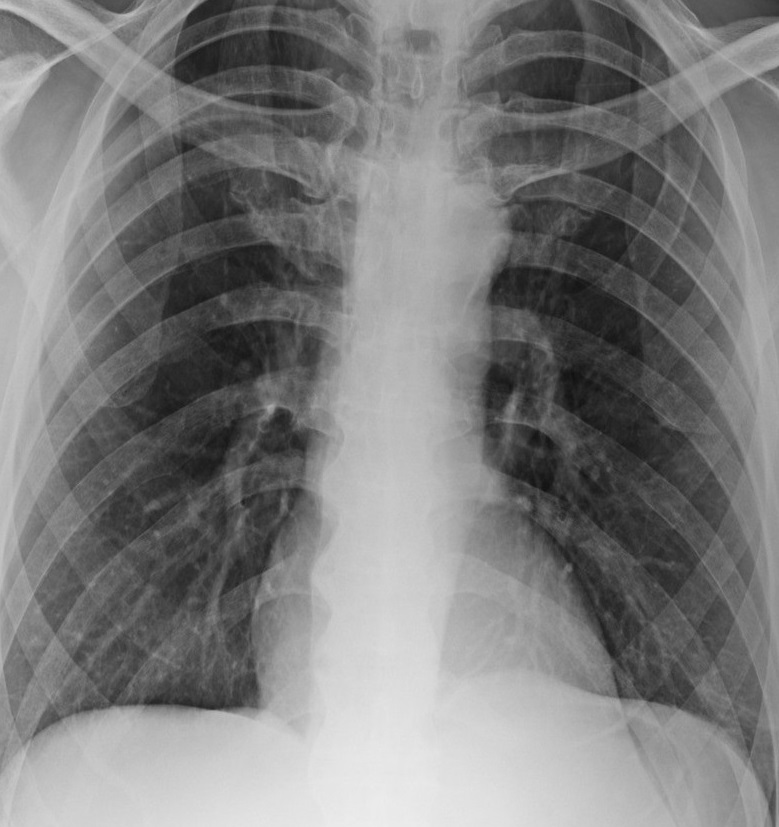}  
  \caption{Normal.}
  \label{fig:sub-non-rydls}
\end{subfigure}
\begin{subfigure}{.22\textwidth}
  \centering
  \includegraphics[width=.9\linewidth]{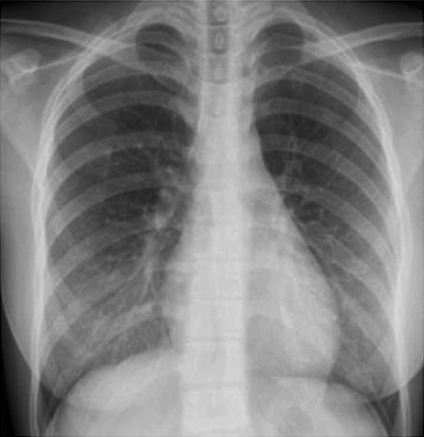}  
  \caption{COVID-19.}
  \label{fig:sub-covid-rydls}
\end{subfigure}
\begin{subfigure}{.22\textwidth}
  \centering
  \includegraphics[width=.9\linewidth]{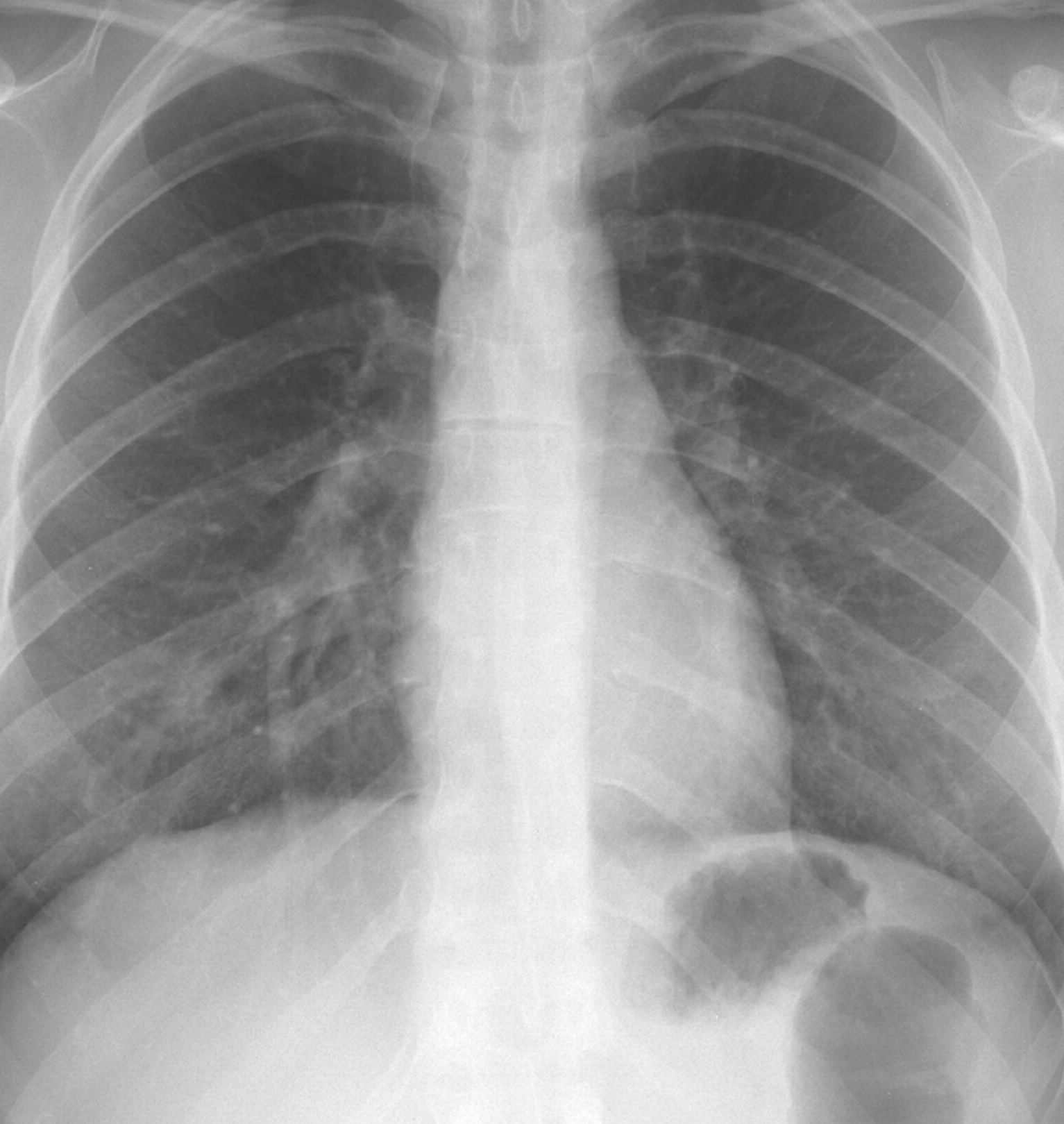}  
  \caption{SARS.}
  \label{fig:sub-sars-rydls}
\end{subfigure}
\begin{subfigure}{.22\textwidth}
  \centering
  \includegraphics[width=.9\linewidth]{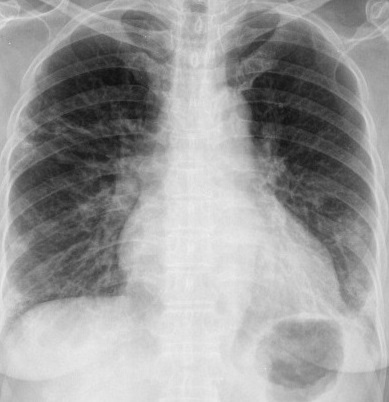}  
  \caption{MERS.}
  \label{fig:sub-mers-rydls}
\end{subfigure}
\newline
\begin{subfigure}{.22\textwidth}
  \centering
  \includegraphics[width=.9\linewidth]{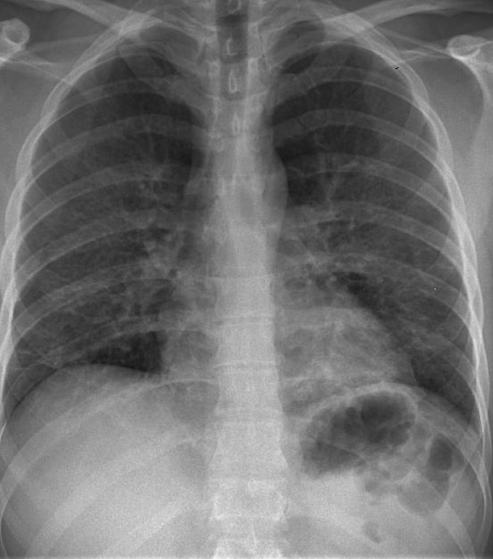} 
  \caption{Pneumocystis.}
  \label{fig:sub-pneumocystis-rydls}
\end{subfigure}
\begin{subfigure}{.22\textwidth}
  \centering
  \includegraphics[width=.9\linewidth]{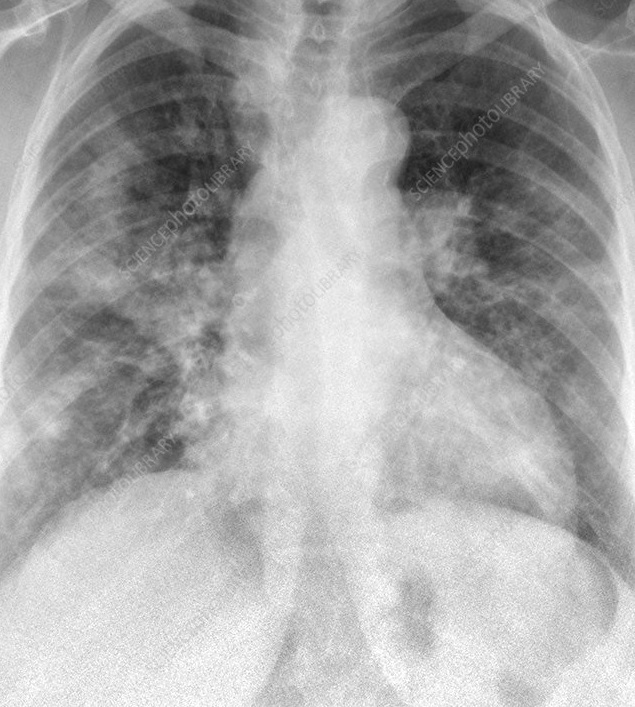}
  \caption{Streptococcus.}
  \label{fig:sub-strepto-rydls}
\end{subfigure}
\begin{subfigure}{.22\textwidth}
  \centering
  \includegraphics[width=.9\linewidth]{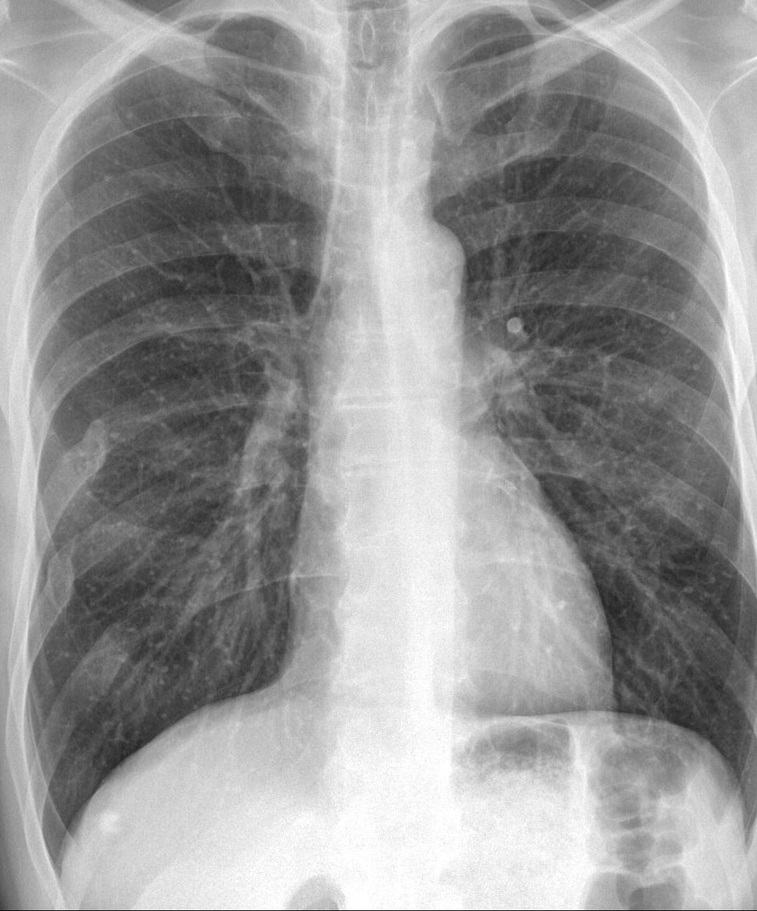}  
  \caption{Varicella.}
  \label{fig:sub-varicella-rydls}
\end{subfigure}
\caption{RYDLS-20 image samples.}
\label{fig:rydls-img}
\end{figure}

Figure \ref{fig:rydls-img} presents image examples for each class retrieved from the RYDLS-20 database. It is worth to mentioning that we have no further information concerning the CXR images with regarding the CXR machine used to take the image, as well as the origin, age and ethnicity of the people whose these images belong to. 

Another worth mentioning information concerning the database is that we have manually cut the images edges in order to avoid the recognition of undesirable patterns. In order to confirm the importance of this preprocessing step, we must cite the work of Maguolo and Nanni (2020) \cite{maguolo2020critic}, in which the authors have made a critical evaluation regarding the combination of different databases for the COVID-19 automatic detection from CXR. In their work, they have cut off the lungs from the x-ray images and have experimentally proved that a classifier can identify from which database the images came from. Thus, the authors highlight that joining different databases may add bias to the classification results, since the classifiers may be recognizing patterns from the origin database and not from the lung injuries. However, as we have manually cut the images edges in RYDLS-20, we have minimized the issue pointed by Maguolo and Nanni (2020) in our experiments.

Table \ref{samp_dist_mc} presents RYDLS-20 samples distribution for the multi-class scenario. Though the table, it is possible to observe the major imbalanceness of the dataset between the pneumonia labels. In most cases, we have seven--nine samples in the training set and only three samples in the test dataset, which makes the learning process much more difficult than in a balanced context. 

\begin{table}[ht]
\centering
\caption{RYDLS-20 samples distribution for the multi-class scenario.}
\begin{tabular}{lccc}
\hline
\textit{Label} & \multicolumn{1}{l}{\textit{\#Samples}} & \multicolumn{1}{l}{\textit{\#Train}} & \multicolumn{1}{l}{\textit{\#Test}} \\ \hline
Normal & 1,000 & 700 & 300 \\
COVID-19 & 90 & 63 & 27 \\
MERS & 10 & 7 & 3 \\
SARS & 11 & 8 & 3 \\
Varicella & 10 & 7 & 3 \\
Streptococcus & 12 & 9 & 3 \\
Pneumocystis & 11 & 8 & 3 \\ \hline
\end{tabular}
\label{samp_dist_mc}
\end{table}

Table \ref{samp_dist_hiear} shows RYDLS-20 samples distribution for the hierarchical classification scenario. This scenario seems less imbalanced because of the internal nodes. For example, all viral pneumonia instances are also categorized as ``Acelluar'', so we have a total of 121 samples labeled with ``Pneumonia/Acellular/Viral''. This less imbalanced context for the hierarchical scenario is partially true, since the hierarchical classifier does use this criteria into account during the learning process.

\begin{table}[htbp]
\centering
\caption{RYDLS-20 samples distribution for the hierarchical scenario.}
\setlength{\tabcolsep}{3.5pt}
\begin{tabular}{lccc}
\hline
\textit{Label Path} & \multicolumn{1}{l}{\textit{\#Samples}} & \multicolumn{1}{l}{\textit{\#Train}} & \multicolumn{1}{l}{\textit{\#Test}} \\ \hline
Normal & 1,000 & 700 & 300 \\
Pneumonia & 144 & 102 & 42 \\
Pneumonia/Acellular & 121 & 85 & 36 \\
Pneumonia/Acellular/Viral & 121 & 85 & 36 \\
Pneumonia/Acellular/Viral/Coronavirus & 111 & 78 & 33 \\
Pneumonia/Acellular/Viral/Coronavirus/COVID-19 & 90 & 63 & 27 \\
Pneumonia/Acellular/Viral/Coronavirus/MERS & 10 & 7 & 3 \\
Pneumonia/Acellular/Viral/Coronavirus/SARS & 11 & 8 & 3 \\
Pneumonia/Acellular/Viral/Varicella & 10 & 7 & 3 \\
Pneumonia/Celullar & 23 & 17 & 6 \\
Pneumonia/Celullar/Bacterial & 12 & 9 & 3 \\
Pneumonia/Celullar/Bacterial/Streptococcus & 12 & 9 & 3 \\
Pneumonia/Celullar/Fungus & 11 & 8 & 3 \\
Pneumonia/Celullar/Fungus/Pneumocystis & 11 & 8 & 3 \\ \hline
\end{tabular}
\label{samp_dist_hiear}
\end{table}

It is important to reinforce the statement that RYDLS-20 labels distribution reflect a real world scenario in which healthy cases are much more frequent (majority class), followed by viral pneumonia (mainly caused by COVID-19), bacterial and fungi pneumonia being the least frequent, in this order.

The experiments were conducted into a split of 70/30 between training and test, which means that we are using the holdout validation technique. It may be asked why do not use a cross-validation schema, since it brings robustness to the experimental results. The answer to this questioning is mainly grounded in the size of the database. As shown in this section, we are dealing with a highly imbalanced database, in which half of the labels has between 10-12 samples. Thus, if we use a 10-fold or 5-fold cross-validation (the most recommended values by the research community), most labels will have only one or two examples into each fold. This division would have a high impact in the testing phase, which may lead to misleading results regarding the evaluation measures. Therefore, the use of a cross-validation technique is conditioned to a bigger database, which is a future work target.

\subsection{Algorithms, Parameters and Metrics}

In this subsection we present the main information concerning the algorithms, parameters and metrics used in the experiments of this work.

\subsubsection{Multi-class Classification}

In order to perform the multi-class classification task we have used five well-known classifiers from the literature: k-Nearest Neighbots (kNN); Support Vectors Machine (SVM); Multilayer Perceptrons (MLP); Dicision Trees (DT); and Random Forests (RF). Table \ref{classic-params} reports the parameter used in these multi-class classification algorithms. 

\begin{table}[ht]
\centering
\caption{Parameter settings of the classic algorithms.}
\label{classic-params}
\begin{tabular}{cll}
\hline
\textit{Algorithm} & \multicolumn{2}{c}{\textit{Parameters}} \\ \hline
\multirow{2}{*}{KNN} & Number of Neighbors & 3 and 5 \\
 & Distance & Euclidean \\ \hline
\multirow{5}{*}{SVM} & Kernel & RBF \\
 & Penalty Parameter (C) & 1 \\
 & Degree & 3 \\
 & Gamma & Scale \\
 & Cache size & 200 \\
 & Decision Function Shape & Ovr \\
 & Tolerance & 0.001 \\ \hline
 \multirow{7}{*}{MLP} & Solver & LBFGS \\
 & Alpha & 1e-5 \\ 
 & Shuffle & True \\ 
 & Max Iterations & 500 \\ 
 & Learning Rate Init & 0.3 \\ 
 & Momentum & 0.2 \\ 
 & Hidden Layer Sizes & 13 \\ \hline
 \multirow{5}{*}{DT} & Criterion & Gini \\ 
 & Splitter & Best \\ 
 & Min Samples Leaf & 10 \\ 
 & Min Samples Split & 20 \\ 
 & Max Leaf Nodes & None \\ 
 & Max Depth & 10 \\ \hline
\multirow{3}{*}{RF} & Number of Trees & 10 \\ 
 & Class Weight & Balance \\ 
 & Type of Trees & Same of DT \\ \hline
\end{tabular}
\end{table}

\subsubsection{Hierarchical Classification}

For the hierarchical classification task we used the Clus-HMC framework\footnote{Available for download at https://dtai.cs.kuleuven.be/clus/}. Clus-HMC was chosen because it is considered in the literature as the state-of-the-art hierarchical classification framework \cite{cerri2015hierarchical, wehrmann2018hierarchical, pereira2019hierarchical}. 

Clus-HMC is based on Predictive Cluster Trees (PCT) and generates a single Decision Tree (DT) considering the entire class hierarchy. In Clus-HMC, DTs are seen as a hierarchy of clusters where the root node contains all the training instances, while the remaining are recursively divided into smaller groups as the hierarchy is traversed towards the leaves. The classification is performed using a distance-based metric which calculates how similar an instance is to some tree. The parameter configurations used in the Clus-HMC algorithm are presented in Table \ref{params}.

\begin{table}[htbp]
\centering
\caption{Clus-HMC execution parameters.}
\setlength{\tabcolsep}{3.5pt}
\begin{tabular}{lc}
\hline
\multicolumn{1}{c}{\textit{Parameter}} & \textit{Value} \\ \hline
Type & Tree \\
ConvertToRules & No \\
HSeparator & ``/''\\
FTest & {[}0.001, 0.005, 0.01, 0.05, 0.1, 0.125{]} \\
EnsembleMethod & RForest \\
Iterations & 10 \\
VotingType & Majority \\
EnsembleRandomDepth & No \\
SplitSampling & None \\
Heuristic & Default \\
PruningMethod & Default \\
CoveringMethod & Standard \\ \hline
\end{tabular}
\label{params}
\end{table}

\subsubsection{Resampling Algorithms}

Regarding the resampling methods, for both classification scenarios (multi-class and hierarchical), we have tested a total of 16 methods, considering oversampling, undersampling and hybrid approaches. However, to avoid visual issues, we report in this work only the algorithms that somehow improved the classification results, which are also the same methods presented in Table \ref{bres_alg} of this work, i.e, ADASYN, SMOTE, SMOTE-B1, SMOTE-B2, AllKNN, ENN, RENN, Tomek Links (TL) and SMOTE+TL.

\subsubsection{Evaluation Metric}

In order to analyze the performance of the experimental results, the F1-Score measure was chosen. Moreover, in order to analyze the general classification performance, we have chosen the macro-avg evaluation, which makes an averaging calculation by class. This is a crucial point in the experimental setup, since evaluation measures such as accuracy may neglect the real performance of the learners for the imbalanced classes, which in our case is the main objective. Following this reasoning, the use of a metric that can really consider the imbalanceness of the different labels is necessary, and, according to Goutte \textit{et al.} \cite{goutte2005probabilistic}, F1-Score is a good alternative to deal with this issue.

As well known in the machine learning community, F1-Score is the harmonic average between precision and recall calculations. Moreover, we have used the macro-avg F1-Score evaluation in order to calculate the mean F1-Score between the classes and not the samples. It is important to observe that we have used the same F1-Score measure in both multi-class and hierarchical classification scenarios.

\section{Experimental Results}
\label{sec:results} 

As we have experimentally tested both multi-class and hierarchical classification scenarios, we have sub-divided the experimental results into two subsections: Multi-Class Classification Results (\ref{subsec:mc_res}) and Hierarchical Classification Results (\ref{subsec:hier_res}). 

We must highlight that in the subsections we present the results considering two perspectives: 
\begin{itemize}
    \item The general macro-avg F1-Score for the evaluated scenarios, i.e., the average F1-Score for all classes in the classification task; and
    \item The F1-Score obtained specifically for the COVID-19 class, given that this is our main interest here. 
\end{itemize}

Due to the large number of experimental results (a total of 3,648 in the Multi-Class scenario and 1,712 in the Hierarchical scenario), we have decided to present only the best result achieved in each prediction schema. However, it is important to mention that, as well as the database and scripts, a complete version of the results is freely available for further analysis\footnote{https://drive.google.com/open?id=1J9I-pPtPfLRGHJ42or4pKO2QASHzLkkj}. Moreover, in order to summarize the results into a unique focal point, in subsection \ref{subsec:sum_res} we also present plots summarizing the best of all results for each classification scenario (Multi-class and Hierarchical). 

\subsection{Multi-Class Results}
\label{subsec:mc_res}

Table \ref{best_covid_mc} shows the best F1-Score results for the COVID-19 identification considering the multi-class scenario. The individual predictions have achieved the same top result as of late fusion method ($\approx$0.83). Moreover, the MLP classifier is present in the best results for all the prediction schemas, as well as LPQ feature, which is also present in all prediction schemas.

\begin{table}[htbp]
\centering
\caption{Best results for COVID-19 label for each prediction schema in the Multi-Class scenario.}
\begin{tabular}{ccccc}
\hline
\textit{Prediction Schema} & \textit{Feature(s)} & \textit{Classifier(s)} & \textit{Resampling(s)} & \textit{F1-Score} \\ \hline
Individual & LPQ & MLP & ENN or None & 0.8333 \\ 
Early Fusion & LBP \& LPQ & MLP & AllKNN or RENN & 0.8000 \\ 
\begin{tabular}[c]{@{}c@{}}Late Fusion\\ (Top-5)\end{tabular} & LPQ & MLP & ENN & 0.8333 \\
\begin{tabular}[c]{@{}c@{}}Late Fusion\\ (Top-Features)\end{tabular} & \begin{tabular}[c]{@{}c@{}}BSIF, EQP\\ \& LPQ\end{tabular} & MLP & ENN \& RENN & 0.8333 \\
\begin{tabular}[c]{@{}c@{}}Late Fusion\\ (Top-Classifiers)\end{tabular} & LDN \& LPQ & MLP \& DT & \begin{tabular}[c]{@{}c@{}}SMOTE+TL\\ \& ENN\end{tabular} & 0.8333 \\ \hline
\end{tabular}
\label{best_covid_mc}
\end{table}

Table \ref{best_macro_mc} shows the macro-avg F1-Score for the prediction schemas in the multi-class scenario. The best result was achieved both with individual and late fusion schemas ($\approx$0.65). Here, the MLP classifier also achieved a good performance, which is present in all schemas with exception of late fusion with top classifiers. However, there is no clear predominance in relation to the best features among these schemas.

\begin{table}[htbp]
\centering
\caption{Best macro-avg results for each prediction schema in the Multi-Class scenario.}
\begin{tabular}{ccccc}
\hline
\textit{Prediction Schema} & \textit{Feature(s)} & \textit{Classifier(s)} & \textit{Resampling(s)} & \textit{F1-Score} \\ \hline
Individual & LBP & MLP & \begin{tabular}[c]{@{}c@{}}RENN or\\ AllKNN\end{tabular} & 0.6491 \\
Early Fusion & \begin{tabular}[c]{@{}c@{}}BSIF \& EQP\\ \& LPQ\end{tabular} & MLP & TomekLink & 0.5563 \\
\begin{tabular}[c]{@{}c@{}}Late Fusion\\ (Top-5)\end{tabular} & LBP & MLP & \begin{tabular}[c]{@{}c@{}}AllKNN \&\\ RENN\end{tabular} & 0.6491 \\
\begin{tabular}[c]{@{}c@{}}Late Fusion\\ (Top-Features)\end{tabular} & BSIF \& LBP & MLP & \begin{tabular}[c]{@{}c@{}}RENN \&\\ SMOTE-B2\end{tabular} & 0.6491 \\
\begin{tabular}[c]{@{}c@{}}Late Fusion\\ (Top-Classifiers)\end{tabular} & \begin{tabular}[c]{@{}c@{}}LDN \&\\ LETRIST\end{tabular} & DT \& KNN-3 & \begin{tabular}[c]{@{}c@{}}RENN\\ or None\end{tabular} & 0.4500 \\ \hline
\end{tabular}
\label{best_macro_mc}
\end{table}

\begin{figure}[hbtp]
    \centering
    \includegraphics[width=0.6\textwidth]{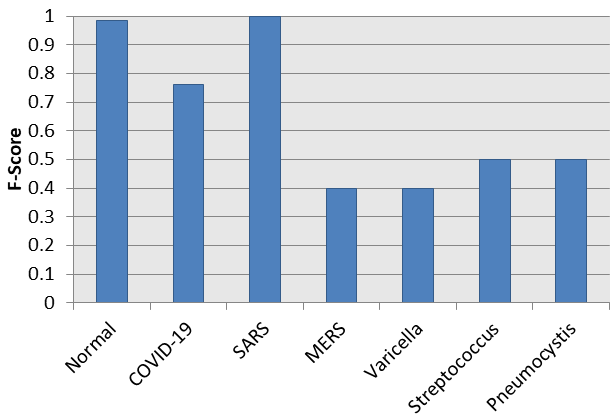}
    \caption{F1-Score results per label in the best case scenario for multi-class context.}
    \label{mc_per_label}
\end{figure}

Figure \ref{mc_per_label} presents a chart of the individual F1-Score results per label for the best case scenario in the multi-class context, which was achieved using the MLP classifier over LBP features after a resampling using the RENN algorithm. The ``SARS'' label and the ``Normal'' label obtained excellent performance with F1-Score 1.0 and 0.98 respectively. Moreover, the COVID-19 label also achieved a satisfactory result ($\approx$0.76). The other labels (MERS, Varicella, Streptococcus and Pneumocystis) reached moderated performances (between 0.4 and 0.5). 

\begin{figure}[hbtp]
    \centering
    \includegraphics[width=0.6\textwidth]{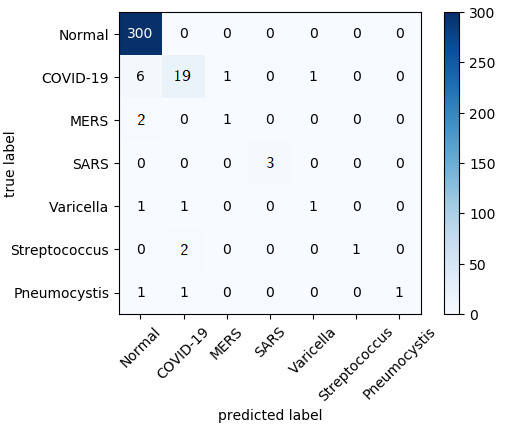}
    \caption{Confusion Matrix in the best case scenario for the multi-class experiments.}
    \label{cm-sl}
\end{figure}

Figure \ref{cm-sl} presents the confusion matrix for the same best case scenario in the multi-class classification context presented in Figure \ref{mc_per_label} (MLP classifier with LBP feature using RENN resampling method).

\subsection{Hierarchical Results}
\label{subsec:hier_res}

Table \ref{best_covid_hier} shows the best F1-Score results for the COVID-19 identification considering the hierarchical scenario. The best F1-Score result for COVID-19 ($\approx$0.89) was achieved in this context. This result was reached using the early fusion prediction schema with the features BSIF, EQP and LPQ resampled by either SMOTE or TL algorithms. Furthermore, the SMOTE resampling algorithm and BSIF features are present in all these classification schemas.

\begin{table}[ht]
\centering
\caption{Best results for COVID-19 label for each prediction schema in the Hierarchical scenario.}
\begin{tabular}{cccc}
\hline
\textit{Prediction Schema} & \textit{Feature(s)} & \textit{Resampling(s)} & \multicolumn{1}{l}{\textit{F1-Score}} \\ \hline
Individual & BSIF & SMOTE-B1 & 0.8387 \\
Early Fusion & BSIF \& EQP \& LPQ & SMOTE or TL & 0.8889 \\
\begin{tabular}[c]{@{}c@{}}Late Fusion\\ (Top-5)\end{tabular} & BSIF \& OBIF & \begin{tabular}[c]{@{}c@{}}SMOTE-B1\\ \& None\end{tabular} & 0.8276 \\
\begin{tabular}[c]{@{}c@{}}Late Fusion\\ (Top-Features)\end{tabular} & BSIF \& EQP & \begin{tabular}[c]{@{}c@{}}SMOTE-B2\\ \& None\end{tabular} & 0.8276 \\ \hline
\end{tabular}
\label{best_covid_hier}
\end{table}

Table \ref{best_macro_hier} presents the macro-avg F1-Score results in the hierarchical scenario. As well as for the COVID-19 identification, the best macro-avg F1-Score result was achieved in the early fusion prediction schema. Furthermore, SMOTE resampling technique also appears in all classification schemas.

\begin{table}[ht]
\centering
\caption{Best macro-avg results for each prediction schema in the Hierarchical scenario.}
\begin{tabular}{cccc}
\hline
\textit{Prediction Schema} & \textit{Feature(s)} & \textit{Resampling(s)} & \multicolumn{1}{l}{\textit{F1-Score}} \\ \hline
Individual & LETRIST & SMOTE-B2 & 0.4615 \\
Early Fusion & \begin{tabular}[c]{@{}c@{}}LBP \& INCEPTION-V3\\ \& LETRIST\end{tabular} & SMOTE-B1 & 0.5669 \\
\begin{tabular}[c]{@{}c@{}}Late Fusion\\ (Top-5)\end{tabular} & LDN \& LETRIST & \begin{tabular}[c]{@{}c@{}}SMOTE \&\\ SMOTE-B2\end{tabular} & 0.4751 \\
\begin{tabular}[c]{@{}c@{}}Late Fusion\\ (Top-Features)\end{tabular} & BSIF \& LETRIST & \begin{tabular}[c]{@{}c@{}}SMOTE-B1 \&\\ SMOTE\end{tabular} & 0.4751 \\ \hline
\end{tabular}
\label{best_macro_hier}
\end{table}

Figure \ref{mc_per_label} presents a chart of the individual F1-Score results per label for the best case scenario in the hierarchical classification schemas, which was achieved in the early fusion schema with LBP, INCEPTION-V3 and LETRIST features after resampling with the SMOTE-B1 algorithm. It is important to observe that, as we are dealing with a hierarchical classification problem, the predictions are made for each label in the hierarchy, thus we have a total of fourteen hierarchically organized labels. We may also observe that the classifier achieved very different performances for each type of pneumonia. While for COVID-19, MERS and Streptococcus, it has achieved a F1-Score close to 0.7, for Varicella and Pneumocystis the F1-Score was zero.

\begin{figure}[hbtp]
    \centering
    \includegraphics[width=0.7\textwidth]{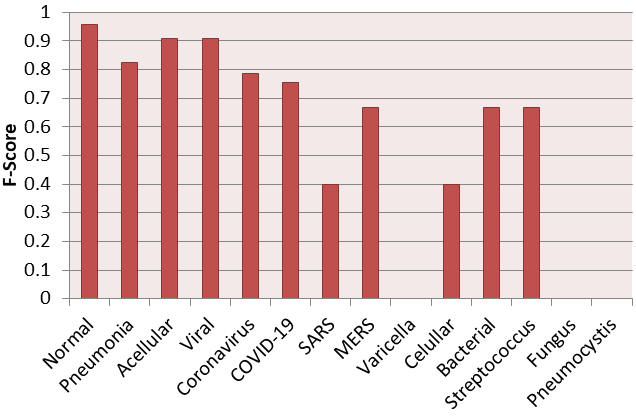}
    \caption{F1-Score results per label in the best case on the hierarchical scenario.}
    \label{hier_per_label}
\end{figure}

\begin{figure}[htbp]
    \centering
    \includegraphics[width=0.9\textwidth]{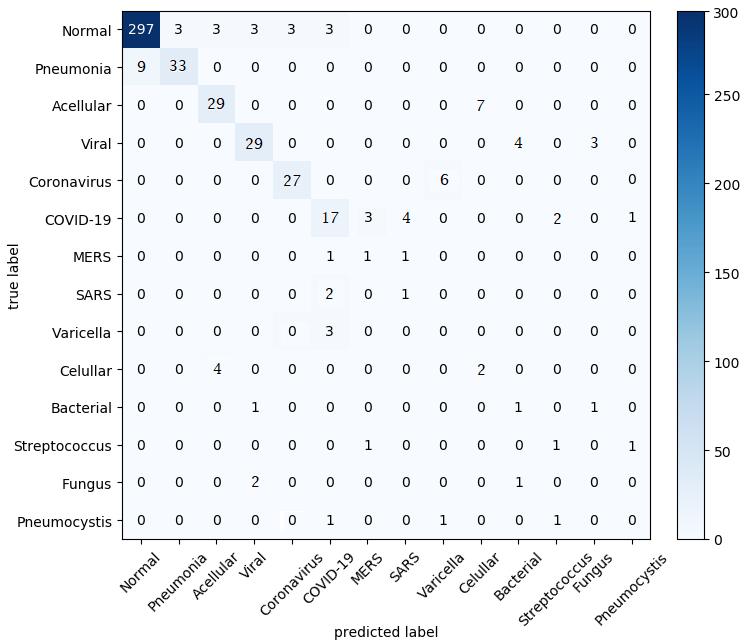}
    \caption{Confusion Matrix in the best case scenario for the hierarchical experiments.}
    \label{cm-h}
\end{figure}

Figure \ref{cm-h} shows the confusion matrix for the same best case scenario in the hierarchical classification context presented in Figure \ref{hier_per_label} (Early Fusion of LBP, INCEPTION and LETRIST using SMOTE-B1 resampling method). It is worth mentioning that as we are dealing with the problem in a hierarchical classification design, the construction of the Confusion Matrix can follow different approaches when compared with the ones built in the multi-class classification scenarios. In order to build the Confusion Matrix presented in Figure \ref{cm-h}, we have followed a method used by Buchanan \textit{et al.} (2019) \cite{buchanan2019towards}, in which we group every label of the hierarchy into the same matrix, i.e., all the internal and leaf label nodes.

\subsection{Results Summary}
\label{subsec:sum_res}

\begin{figure}[htbp]
    \centering
    \includegraphics[width=0.6\textwidth]{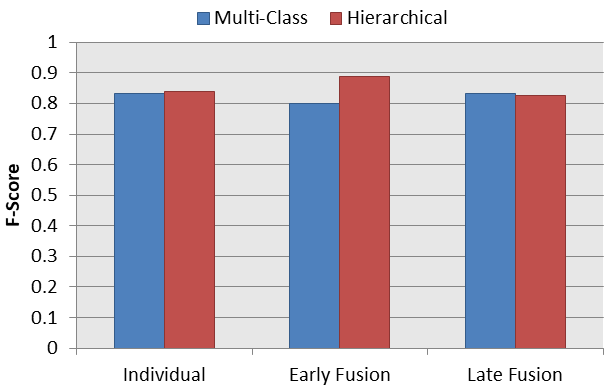}
    \caption{Best F1-Score results on multi-class and hierarchical scenarios for COVID-19 Identification.}
    \label{summary_covid}
\end{figure}

Figure \ref{summary_covid} presents a summary chart of the best F1-Score results for the COVID-19 identification in the three classification schemas (individual, early fusion and late fusion) both in multi-class and hierarchical classification scenarios. The hierarchical classification outperformed the multi-class classification in the individual and early fusion schemas. In addition, the best of all COVID-19 identification result was achieved by the hierarchical classification learner.

\begin{figure}[htbp]
    \centering
    \includegraphics[width=0.6\textwidth]{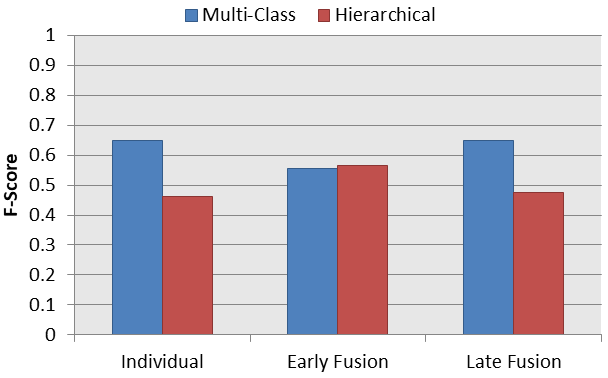}
    \caption{Best macro-avg F1-Score results on multi-class and hierarchical scenarios.}
    \label{summary_macro}
\end{figure}

Figure \ref{summary_macro} presents a summary of the best macro-avg F1-Score results. We may note that in two (i.e. individual and late fusion) out of the three schemas, the best result was achieved by far in the multi-class classification scenario. Moreover, the best of all macro-avg F1-Score ($\approx$0.65) was performed by the multi-class learners in the individual and late fusion schemas.

\section{Discussions}
\label{sec:discussion}

Aiming to evaluate the obtained results from different points of view, we guide our discussion in the search for answers to the following questions:
\begin{itemize}
\item Which feature representation provided the best results?
\item Which base classifier performed better in the multi-class scenario?
\item Which resampling algorithms improved the results the most?
\item Have the fusion strategies contributed to improve the results?
\item Which kinds of labels are easier/harder to predict?
\item Which labels were mixed up in the best case scenario for each classification schema?
\item What is the impact of the type of classification in the results for this domain scenario (multi-class \textit{versus} hierarchical)?
\item What may be happening in the misclassified CXR?
\end{itemize}

In the following subsections, we answer each question taking into account the statistical significance, when it is suitable.

\subsection{Which feature representation provided the best results?}

To answer this question, i.e., define the best feature representation, we have used the statistical protocol proposed in Charte \textit{et al.} \cite{charte2015mlsmote}. With this protocol, we calculate the ranking of the feature in each classification scenario based on the Friedman statistical test. In other words, the features performance are ranked (from first to last) and an average rank is calculated for each classification scenario. As we have four different measures contexts, i.e., the macro-avg of the results and F1-Score for the COVID-19 in the multi-class and hierarchical scenarios, we have performed this protocol four times, one in each classification context and measure. The test results in relation to the ranking of the features for the COVID-19 Identification and the macro-avg F1-Score are shown in Table \ref{fried_features}. It is worth mentioning that, since this test is composed of rankings, which ranges from the first to the last, the lower the ranking score is, the better is the performance.

We can observe that, in the multi-class scenario for COVID-19 identification, LPQ achieved the best average ranking by far, being in first place for all scenarios. In the hierarchical context, BSIF feature achieved the best ranking for the COVID-19 identification also with an unanimous first place. Considering the macro-avg results in the multi-class scenario, the best ranked feature in average was LBP (with an average ranking of 1.33), while in the hierarchical scenario it was LETRIST also with an unanimous first place. Moreoever, analyzing the overall average ranking, we have LPQ as the best feature considering all classification contexts and scenarios with an overall ranking of 2.58.

\begin{table}[htbp]
\centering
\caption{Ranking of the results per feature set in all classification scenarios.}
\scalefont{0.9}
\setlength{\tabcolsep}{3pt}
\begin{tabular}{rccccc}
\cline{2-6}
\multicolumn{1}{l}{} & \multicolumn{2}{c}{\textit{Multi-Class}} & \multicolumn{2}{c}{\textit{Hierarchical}} & \multirow{2}{*}{\textit{\begin{tabular}[c]{@{}c@{}}Overall\\ Avg. Ranking\end{tabular}}} \\ \cline{1-5}
\multicolumn{1}{c}{\textit{Feature Set}} & \textit{COVID-19} & \textit{Macro-Avg} & \textit{COVID-19} & \textit{Macro-Avg} &  \\ \hline
BSIF & 4.00 & 3.67 & 1.00 & 3.00 & 2.92 \\
INCEPTION-V3 & 6.33 & 5.00 & 4.67 & 3.33 & 4.83 \\
LBP & 4.67 & 1.33 & 4.00 & 3.00 & 3.25 \\
LDN & 3.00 & 4.33 & 4.00 & 3.00 & 3.58 \\
LETRIST & 5.33 & 2.33 & 2.67 & 1.00 & 2.83 \\
EQP & 4.00 & 5.00 & 3.33 & 4.67 & 4.25 \\
LPQ & 1.00 & 3.33 & 3.00 & 3.00 & 2.58 \\
OBIF & 3.33 & 4.33 & 2.33 & 3.00 & 3.25 \\ \hline
\end{tabular}
\label{fried_features}
\end{table}

\subsection{Which base classifier performed better as a whole in the multi-class scenario?}

In order to answer this question, we have used the same statistical protocol used in the first question, but ranking the base classifiers instead of features. Thus, in Table \ref{fried_class} we present the average ranking per classifier for the COVID-19 identification and the macro-avg F1-Score, both for the multi-class scenario. It can be noted that, regardless the classification context, MLP was by far the best ranked classifier with the lower average overall ranking (1.17). This MLP dominance can be explained by the fact that it can learn intrinsic characteristics of the problem when given the correct amount of features, which should be happening in our application domain.

\begin{table}[htbp]
\centering
\caption{Ranking of results per classifier in the multi-class classification scenario.}
\begin{tabular}{rccc}
\hline
\multicolumn{1}{c}{\textit{Classifier}} & \textit{COVID-19} & \textit{Macro-Avg} & \textit{\begin{tabular}[c]{@{}c@{}}Overall \\ Avg. Rank.\end{tabular}} \\ \hline
KNN & 4.67 & 1.67 & 3.17 \\
RF & 3.00 & 4.67 & 3.83 \\
SVM & 3.67 & 4.33 & 4.00 \\
DT & 2.67 & 3.00 & 2.83 \\
MLP & 1.00 & 1.33 & 1.17 \\ \hline
\end{tabular}
\label{fried_class}
\end{table}

\subsection{Which resampling algorithms improved the results the most?}

For this answer, we have used again the same statistical protocol used in the first two questions. However, in this case, ranking the resampling methods. Table \ref{fried_resamp} presents the results for this test in all the classification contexts. We may observe that, in the multi-class scenario, ENN resampling method was the most effective in improving the COVID-19 identification (with an average ranking of 1.67), while for the macro-avg F1-Score, the most effective resampling methods were AllKNN and RENN, average rankings of 2.33. In the hierarchical context, the resampling technique which, in average, improved the COVID-19 identification the most was SMOTE, with an average ranking of 2.67. Moreover, for the macro-avg F1-Score, SMOTE and SMOTE Borderline-1 resampling methods were the most effective ones, with average rankings of 1.67. 

\begin{table}[htbp]
\centering
\caption{Ranking of results per resampling method in the classification scenarios.}
\setlength{\tabcolsep}{3pt}
\begin{tabular}{rccccc}
\cline{2-6}
\multicolumn{1}{l}{\textit{}} & \multicolumn{2}{c}{\textit{Multi-Class}} & \multicolumn{2}{c}{\textit{Hierarchical}} & \multirow{2}{*}{\textit{\begin{tabular}[c]{@{}c@{}}Overall\\ Avg. Rank.\end{tabular}}} \\ \cline{1-5}
\multicolumn{1}{c}{\textit{Resampling}} & \textit{COVID-19} & \textit{Macro-Avg} & \textit{COVID-19} & \textit{Macro-Avg} &  \\ \hline
ADASYN & 7.67 & 7.67 & 4.67 & 4.00 & 6.00 \\
AllKNN & 4.33 & 2.33 & 5.33 & 7.33 & 4.83 \\
ENN & 1.67 & 4.00 & 5.33 & 6.33 & 4.33 \\
RENN & 3.00 & 2.33 & 6.33 & 6.67 & 4.58 \\
SMOTE & 6.33 & 5.00 & 2.67 & 1.67 & 3.92 \\
SMOTE-B1 & 5.67 & 5.00 & 3.67 & 1.67 & 4.00 \\
SMOTE-B2 & 7.00 & 4.67 & 4.33 & 2.00 & 4.50 \\
SMOTE+TL & 6.33 & 7.33 & 4.00 & 3.00 & 5.17 \\
TL & 3.00 & 3.33 & 3.67 & 5.33 & 3.83 \\ \hline
\end{tabular}
\label{fried_resamp}
\end{table}

Another interesting point to observe in the test results shown in Table \ref{fried_resamp} is that in an overall average ranking vision, considering all classification scenarios, Tomek Link (TL) technique was the most effective (ranking of 3.83). This result can be explained by the fact that TL is an undersampling method aimed in the removal of pair of instances that can cause confusion in the learner, since they are labaled with different classes but have similar attribute values. Given the domain of application being investigated in this work, this is the exact case that can cause drawbacks in the classification results, since some CXR images are very alike but contains different types of pneumonia.

\subsection{Have the fusion strategies contributed to improve the results?}

Yes, the early fusion strategy was particularly important to provide better results for the COVID-19 identification in the hierarchical scenario. Besides, the early fusion technique also provided a better macro-avg F1-Score in this same classification scenario. However, the late fusion was not able to increase the classification results.

On the other hand, in the multi-class scenario, for the COVID-19 identification, the best results were obtained both with an individual classifier and using the late fusion technique. The same occurred for macro-avg F1-Score. Thus, in the multi-class classification scenario, the fusion was not effective in none of the strategies (early nor late).

\subsection{Which kinds of labels are easier/harder to predict?}

In the best case of macro-avg prediction for the multi-class classification scenario, Normal, COVID-19 and SARS labels presented a remarkably better performance compared to the other labels.

The reason why Normal and COVID-19 performed better leans on the fact that these are the two biggest classes in terms of the number of samples in the database. So, the classifiers must have learned more from their characteristics. Regarding the SARS label, the MLP classifier probably obtained a good performance because it could extract the peculiar visual content of the images, which has a general appearance less dark than the images of the other labels.

In the hierarchical classification context, we have much more labels, since the internal nodes are also considered during the prediction process. In this scenario, we had excellent/good performances for Normal, Pneumonia, Acellular, Viral, Coronavirus and COVID-19 labels. The same justification used in the multi-class scenario can be ruled here, since we have much more samples belonging to the Normal and COVID-19 classes. In addition, the Pneumonia, Acellular, Viral and Coronavirus labels are also assigned in the COVID-19 label path during the prediction. The negative point in the hierarchical classification is related to the Varicella and Pneumocystis labels, which had zero F1-Scores in the best macro-avg scenario. This bad performance may be caused by a generalization issue in the classifier, which is made in order to obtain a better overall macro-avg result, practically ignoring some difficult minority classes.

\subsection{Which labels were mixed up in the best case scenario for each classification schema?}

The answer to this question is grounded in the confusion matrices presented in Section 6 (Experimental Results). Looking at the confusion matrix for the multi-class classification scenario (Figure \ref{cm-sl}), we may observe that six ``COVID-19'' samples were mixed up with the ``Normal'' label. Moreover, two ``Streptococcus'' samples, one ``Varicella'' and one ``Pneumocystis'' were mixed up with the ``COVID-19'' label. Although we do not have a mixed up pattern, it is interesting to observe that samples from different labels were mixed up with the ``Normal'' label.

On the other hand, looking at the confusion matrix for the hierarchical classification scenario (Figure \ref{cm-h}), we may observe more details concerning the classifiers mixed ups, since we have internal labels nodes besides the leaf nodes. In this scenario, we may observe that nine samples with the ``Pneumonia'' label were misclassified as ``Normal''. Moreover, we can also note that ``COVID-19'' was mixed up with all the other pathogens in a certain level, since three ``COVID-19'' samples were misclassified as ``MERS'', four as ``SARS'', two as ``Streptococcus'' and one as ``Pneumocystis''. Concerning the mixed up between the pneumonia pathogens in a high level, we can observe that four ``Viral'' samples were mixed up with ``Bacterial'' and three with ``Fungus''.

\subsection{What is the impact of the type of classification in the results for this application domain (multi-class \textit{versus} hierarchical)?}

Looking at the result tables from Section \ref{sec:results} we can clearly see a difference between the performances in the multi-class and hierarchical classification scenarios. At a first glance, we may say that of course this difference occurs because of the very different classifiers being applied in each scenario. Nevertheless, we may go beyond in this analysis. Even though the distinction between viral and bacterial pneumonia is not relevant enough for the clinical workflows, the hierarchical representation of the problem, proposed here, has shown to be a feasible way to improve the performance on COVID-19 identification. As we have a hierarchical classifier (Clus-HMC) being applied to the problem, it was able to learn relevant information from the hierarchy in order to distinguish some types of pneumonia and maximize exactly the label of greatest interest: COVID-19.

\subsection{What may be happening in the misclassified CXR?}

\begin{figure}[htbp]
\centering
\begin{subfigure}{.24\textwidth}
  \centering
  \includegraphics[width=.9\linewidth]{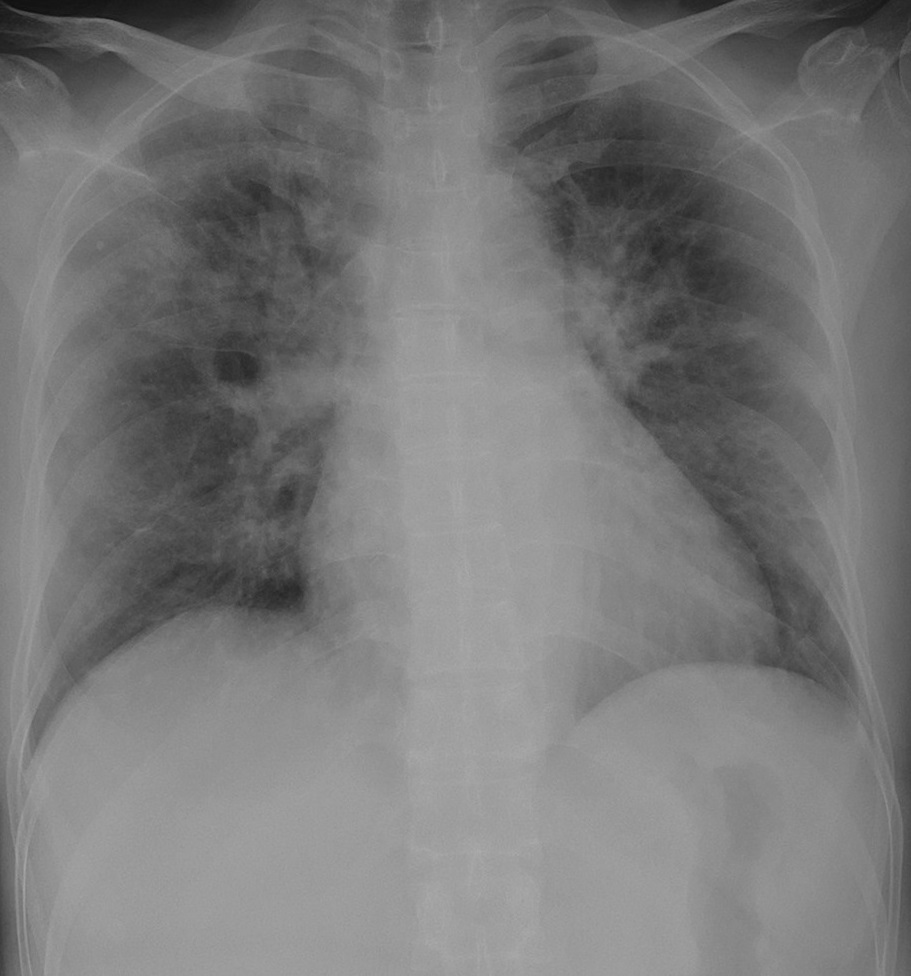} 
  \caption{Example 1.}
  \label{ex_covid_1}
\end{subfigure}
\begin{subfigure}{.24\textwidth}
  \centering
  \includegraphics[width=.9\linewidth]{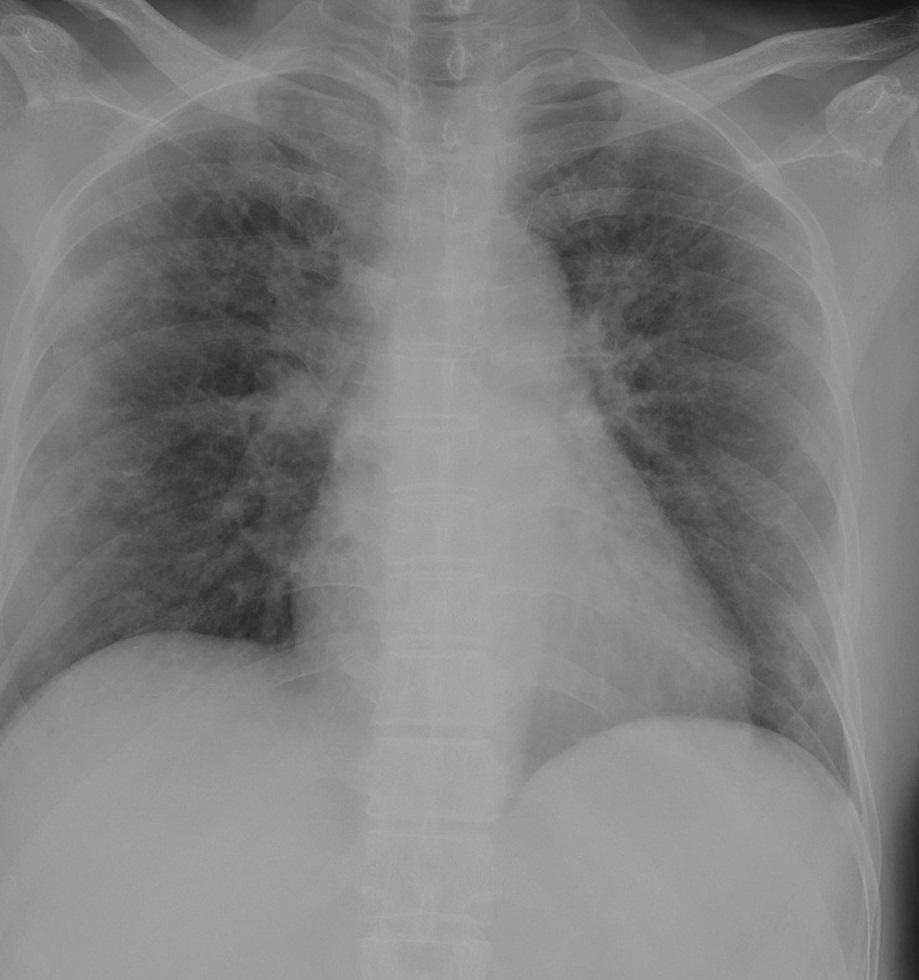}
  \caption{Example 2.}
  \label{ex_covid_2}
\end{subfigure}
\caption{Examples of samples with ``COVID-19'' label that were predicted as ``Normal''.}
\label{fig:ex_covid_normal}
\end{figure}

This is probably the most tricky question in this discussion so far, since CXR images are not always the medical standard to diagnose pneumonia pathogens. Figure \ref{fig:ex_covid_normal} presents two examples of samples with the ``COVID-19'' label that were misclassified as ``Normal'' in the best case scenario of the multi-class classification approach. In these examples, it is really difficult to identify what could make the learner recognize ``Normal'' patterns instead of ``COVID-19''. However, in the following we present other examples that can bring some thoughts into the light of this issue.

\begin{figure}[htbp]
\centering
\begin{subfigure}{.24\textwidth}
  \centering
  \includegraphics[width=.9\linewidth]{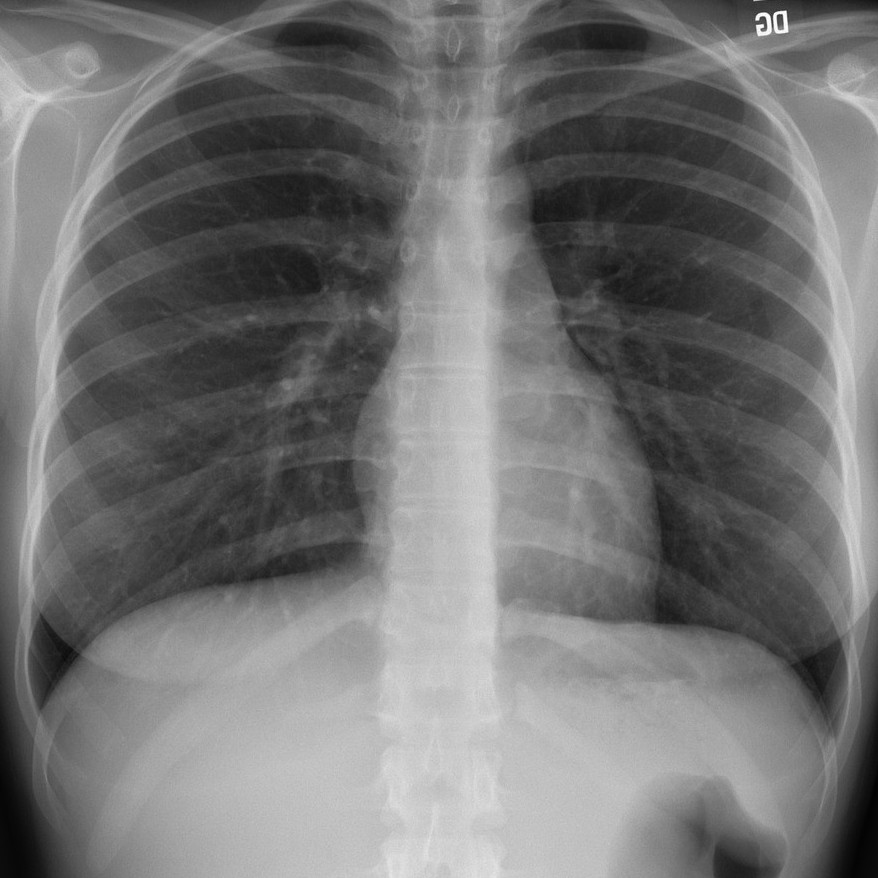}  
  \caption{Normal Example 1.}
  \label{ex_normal1}
\end{subfigure}
\begin{subfigure}{.24\textwidth}
  \centering
  \includegraphics[width=.9\linewidth]{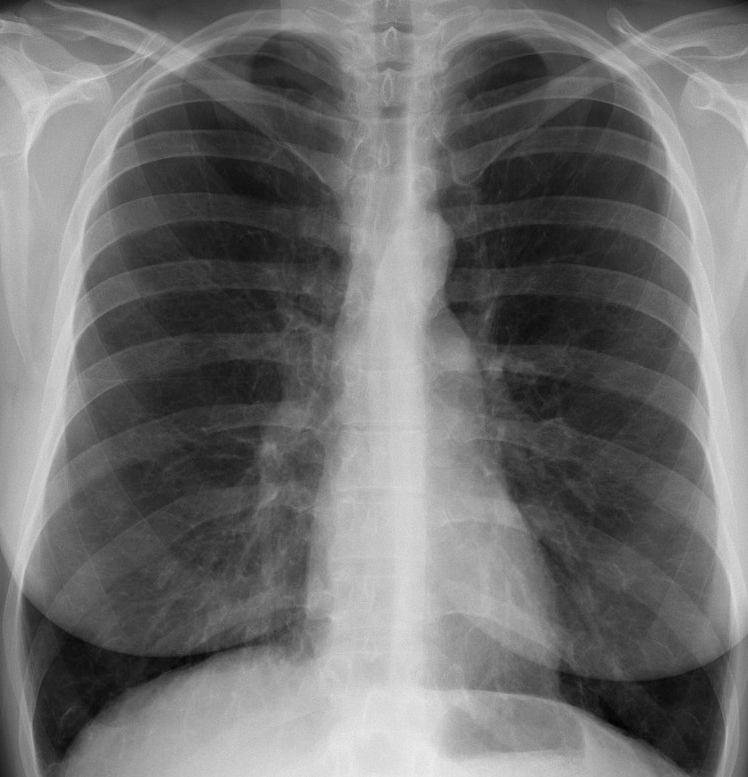}  
  \caption{Normal Example 2.}
  \label{ex_normal2}
\end{subfigure}
\begin{subfigure}{.24\textwidth}
  \centering
  \includegraphics[width=.9\linewidth]{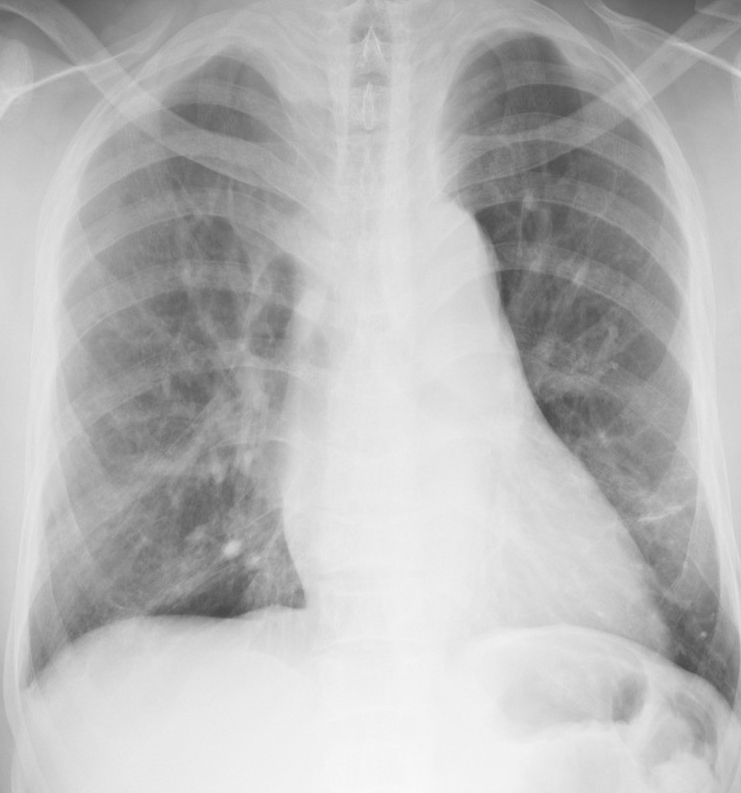} 
  \caption{Normal Example 3.}
  \label{ex_normal3}
\end{subfigure}
\begin{subfigure}{.24\textwidth}
  \centering
  \includegraphics[width=.9\linewidth]{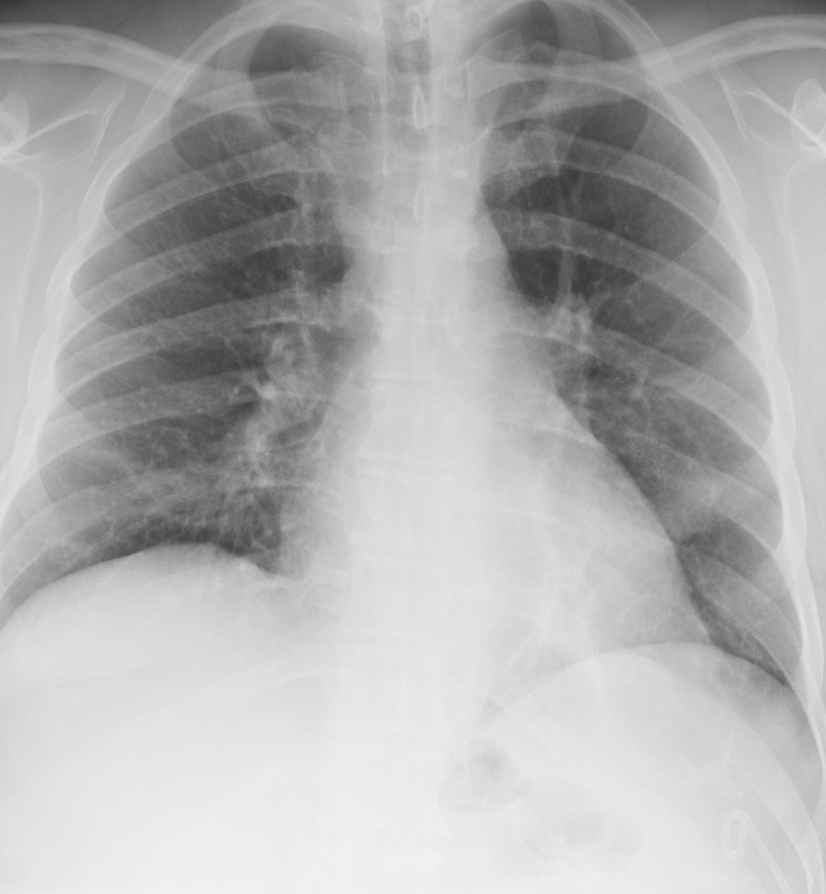}
  \caption{Normal Example 4.}
  \label{ex_normal4}
\end{subfigure}
\caption{Different examples of CXR with ``normal'' lungs.}
\label{fig:ex_cxr}
\end{figure}

In order to give a direction regarding what may be happening in some of the misclassified CXR cases, we present in Figure \ref{fig:ex_cxr} four examples of CXR images from ``normal'' lungs, which were also extracted from RYDLS-20.

When we think about a ``normal'' CXR, i.e., a CXR of a person without pneumonia, we may immediately think in a CXR similar to Figures \ref{ex_normal1} and \ref{ex_normal2}, in which the image is pretty clear and we may see that there are no white spots whatsoever. However, in practice, due to variate factors such as the patient lung characteristics, type of x-ray scan machine, and even due to the protocol followed by the professional radiologist which operates the scan machine, there might be visual variations between the x-rays captured in this different circumstances that still belongs to the same class. Considering this context, it is common to observe in the database samples from people without pneumonia but with CXR images similar to the ones presented in Figures \ref{ex_normal3} and \ref{ex_normal4}. Thus, given as example the four samples from Figure \ref{fig:ex_cxr}, which belong to the same class but present such different characteristics, it is comprehensive that an artificial intelligence system, which is based on similarity patterns, can make recognition mistakes for these images.

\section{Concluding Remarks and Future Works}
\label{sec:conclusions}

As the COVID-19 pandemic spreads around the world, the number of cases keeps growing exponentially. Finding a method that can help in the diagnosis of this disease in people, using a cheap and fast method, is fundamental to avoid overwhelming the healthcare system. In this context, the use of machine learning techniques to identify the pneumonia disease in CXR images has been proposed in the literature and may help in this diagnosis. However, when we are dealing with images taken from patients stricken of pneumonia caused by different types of pathogens and we are trying to predict a specific type of pneumonia (in this case, COVID-19), the problem turns into an even more challenging task.

In a real world context, we have many more people unaffected by pneumonia than affected. In addition, there is a natural imbalance between the number of people stricken of pneumonia caused by different pathogens and, furthermore, it is each time harder to estimate the precise imbalance between these numbers, due to the COVID-19 outbreak. Considering a realistic scenario, in this paper we have proposed a classification schema, aiming to classify pneumonia disease caused by different pathogens in CXR images, and also to identify COVID-19 among them. In the proposed schema we use resampling techniques in order to deal with the intrinsic imbalanceness issue of the problem. Moreover, the proposed schema is composed of eight different feature sets, which are extracted from the images and tested as individual and combined in an early fusion design. Besides, the prediction outputs are also tested individually and in a late fusion design. The proposed schema also foresee the use of flat (multi-class) and hierarchical classifiers. In order to apply hierarchical classification into this application domain, we have considered a tree taxonomy in which the pneumonia are hierarchically organized.

In order to test the proposed classification schema, we have also proposed the RYDLS-20 database. The database is composed of 1,144 CXR images from seven classes: Normal lungs and lungs affected by COVID-19, MERS, SARS, Varicella, Streptococcus and Pneumocystis. The dataset is highly imbalanced, with 1,000 images being from people unaffected by pneumonia, 90 of people affected by COVID-19 and the rest almost equally divided among the other pathogens. The CXR images were obtained from three different sources: Dr. Josepth Cohen GitHub repository; Radiopedia; and NIH dataset.

The proposed classification schema achieved a macro-avg F1-Score of 0.65 using a multi-class approach with MLP classifier using the LBP feature set and resampled with ENN. Furthermore, the proposed schema was also able to achieve a F1-Score of 0.89 for the COVID-19 identification in the hierarchical classification scenario using the early fusion combination of BSIF, EQP and LPQ features resampled with SMOTE+TL. 

It is not fair to make a direct comparison of identification rates obtained in out work and others from the literature, as they are not even necessarily evaluated on the same dataset and under the same circumstances. However, we can remark that, to the best of our knowledge, the best identification rate obtained here (i.e. 0.8889 of F1-Score) is the best nominal rate ever obtained for the task of COVID-19 identification in an unbalanced environment with more than three classes. Moreover, we must highlight the novel proposed hierarchical classification approach considering different types of pneumonia, which lead us to the best recognition rate for COVID-19 in this work.

Even though this proposal does not provide a definitive COVID-19 diagnosis, and this is not the purpose of this work, the good identification rate achieved for COVID-19 can be quite useful to help the screening of patients in the emergency medical support services, that has been severely affected by the pandemic breakthrough. The main virtue of this work is to point out a promising way, although we know that new results must be investigated with new and more robust databases.

As future work, we hope to build a larger database so we can apply more sophisticated Deep Learning techniques with proper depth into the samples. Besides, with a larger database, we can also test our proposed schema in a bigger scale and with a cross-validation approach, providing a robust vision of our proposal into the problem. Moreover, other approaches can be used in the hierarchical classification task, such as the use of Local Classifiers instead of a Global Classifier. The extraction of other feature sets may be also experimentally tested into our proposed classification schema.

\section*{Acknowledgments}
We thank the Brazilian Research Support Agencies: Coordination for the Improvement of Higher Education Personnel (CAPES), National Council for Scientific and Technological Development (CNPq), and Araucaria Foundation (FA) for their financial support. We also thank Dr. Joseph Paul Cohen from the University of Montreal for providing such a useful dataset of pneumonia images for the research community.

\bibliographystyle{unsrt}  
\bibliography{references.bib}

\end{document}